%% file: main.tex
\documentclass[letterpaper, 10 pt, journal, twoside]{IEEEtran}
\IEEEoverridecommandlockouts

\usepackage{decar-common}
\usepackage{decar-dynamics}
\usepackage{decar-lie}
\usepackage{decar-post}

\newlength{\subfigheight}
\newsavebox{\subfigbox}

\makeatletter
\AtBeginDocument{
  \check@mathfonts
}
\newcommand{\partsmash}[2][tb]{%
  \def\mb@t{\ht\z@ #2\ht\z@}\def\mb@b{\dp\z@ #2\dp\z@}%
  \def\mb@tb{\mb@t \mb@b}%
  \edef\finsm@sh{\csname mb@#1\endcsname\box\z@}%
  \ifmmode \@xp\mathpalette\@xp\mathsm@sh
  \else \@xp\makesm@sh
  \fi}
\makeatother

\newcommand{\expectq}[1]{\mathrm{E}_q[#1]}

\usepackage{xfrac}                  
\usepackage{tabularx}               
\usepackage{booktabs}               
\DeclareMathOperator{\meter}{m}

\Crefformat{figure}{#2Fig.~#1#3}

\definecolor{dark_grey}{rgb}{0.45, 0.45, 0.45}
\definecolor{light_grey}{rgb}{0.75, 0.75, 0.75}

\makeatletter
\g@addto@macro\normalsize{%
  \setlength\abovedisplayskip{5pt}
  \setlength\belowdisplayskip{5pt}
  \setlength\abovedisplayshortskip{3pt}
  \setlength\belowdisplayshortskip{3pt}
}
\makeatother

\newcommand{\update}[1]{\colour{black}{#1}}

\newcommand{\secondupdate}[1]{\colour{black}{#1}}

\newcommand{\thirdupdate}[1]{\colour{black}{#1}}

\begin{document}

\input{z_arxiv-cover-ieee.tex}

\fontdimen16\textfont2=\fontdimen17\textfont2
\fontdimen13\textfont2=5pt

\title{Mind the Gap: Norm-Aware Adaptive Robust Loss for Multivariate Least-Squares Problems}

\author{Thomas~Hitchcox and~James~Richard~Forbes%

\thanks{Manuscript received: February 16, 2022; Revised: April 22, 2022;
Accepted: May 16, 2022. This paper was recommended for publication by Editor
Lucia Pallottino upon evaluation of the Associate Editor and Reviewers'
comments.  This work was supported by Voyis Imaging Inc. through the Natural
Sciences and Engineering Research Council of Canada (NSERC) Collaborative
Research and Development (CRD) program, and the McGill Engineering Doctoral
Award (MEDA) program.}

\thanks{Thomas~Hitchcox and James~Richard~Forbes are with the Department of Mechanical
Engineering, McGill University, Montreal, Quebec H3A~0C3, Canada.
\texttt{thomas.hitchcox@mail.mcgill.ca, james.richard.forbes@mcgill.ca}.}%

\thanks{Digital Object Identifier (DOI): see top of this page.}}%

\markboth{IEEE Robotics and Automation Letters. Preprint Version. Accepted May,
2022}{Hitchcox and Forbes: Mind the Gap: Norm-Aware
Adaptive Robust Loss for Multivariate Least-Squares Problems}

\maketitle

\begin{abstract}
    Measurement outliers are unavoidable when solving real-world robot state
    estimation problems.  A large family of robust loss functions (RLFs) exists
    to mitigate the effects of outliers, including newly developed adaptive
    methods that do not require parameter tuning.  All of these methods assume
    that residuals follow a zero-mean Gaussian-like distribution.  However, in
    multivariate problems the residual is often defined as a norm, and norms
    follow a Chi-like distribution with a non-zero mode value.  This produces a
    ``mode gap'' that impacts the convergence rate and accuracy of existing
    RLFs.  The proposed approach, ``Adaptive MB,'' accounts for this gap by
    first estimating the mode of the residuals using an adaptive Chi-like
    distribution.  Applying an existing adaptive weighting scheme only to
    residuals greater than the mode leads to more robust performance and faster
    convergence times in two fundamental state estimation problems, point cloud
    alignment and pose averaging. 
\end{abstract}


\section{Introduction}
\label{sec:intro}

\begin{figure}[htb]
	\centering
	\includegraphics[width=\columnwidth]{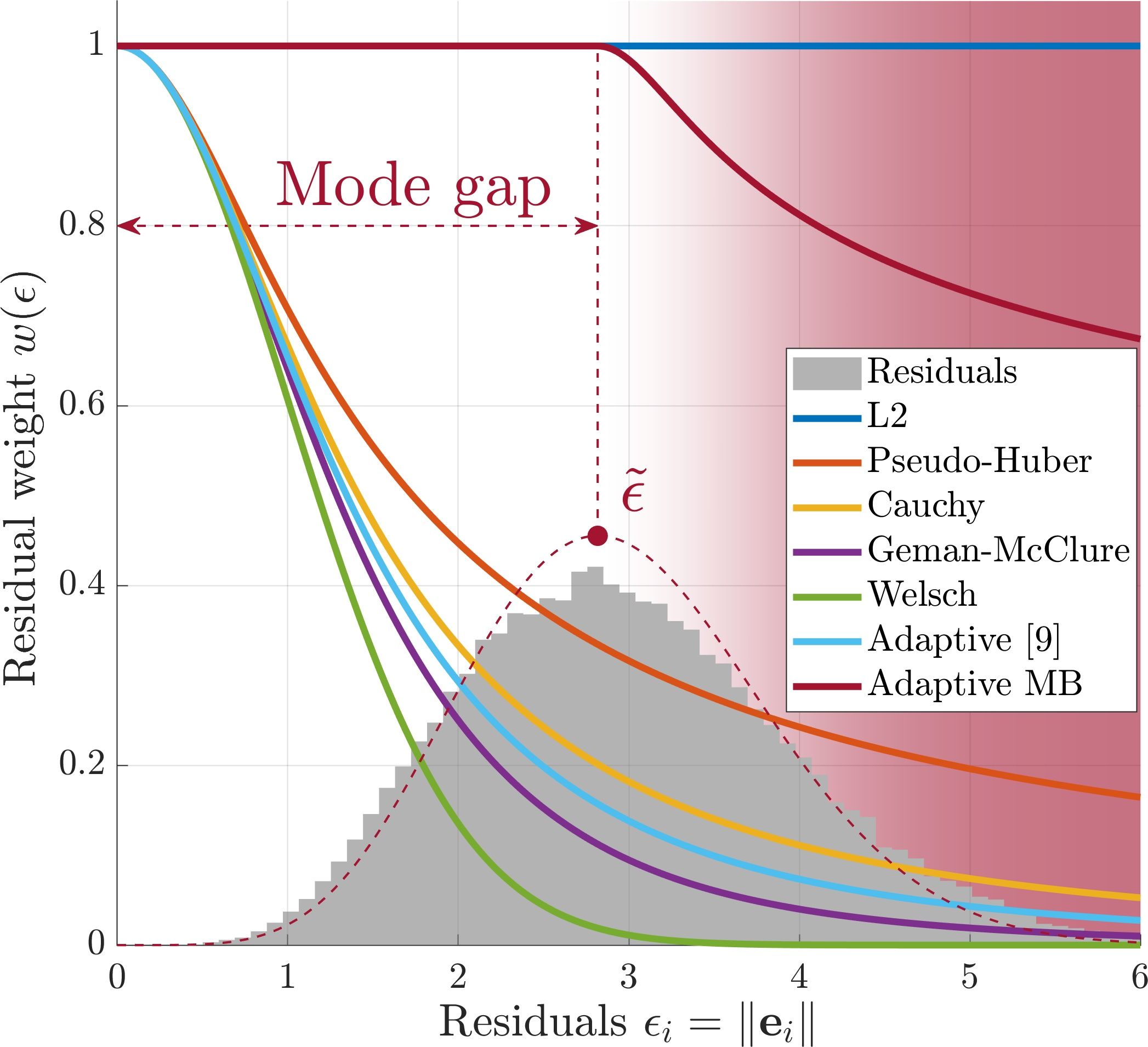}
	\caption{In multivariate least-squares problems, residuals are often defined
	as norms.  If $\mbf{e}_i$ is normally distributed, residual ${\epsilon_i =
	\| \mbf{e}_i \|}$ will follow a Chi distribution with mode
	$\tilde{\epsilon}$.  The proposed robust loss function, ``Adaptive MB,''
	accounts for this ``mode gap'' when assigning residual weights.  As a
	result, ${w_i < 1}$ only in the outlier region (red), leading to more robust
	performance and faster convergence when outliers are present. }
	\label{fig:showcase}
\end{figure}

\IEEEPARstart{M}{easurement} outliers occur in state estimation problems due to
erroneous sensor measurements \cite{Roysdon2017} or faulty data association
decisions \cite{Neira2001}.  Outliers degrade state-estimation solution
accuracy, as residuals formed from outlier measurements can have an outsized
impact when attempting to minimize an objective function.  To address this, a
large family of \textit{robust loss functions} (RLFs) may be used to modify the
objective function in a way that downweights residuals caused by outliers.
Popular RLFs in the literature include pseudo-Huber (also referred to as L1-L2)
\cite{Huber1964,Zhang1997}, Cauchy \cite{Black1996}, Geman-McClure
\cite{Geman1985}, and Welsch \cite{Dennis1978} loss, with each RLF defining a
distribution on the residual weights (see \Cref{fig:showcase}).  

The drawback to these conventional RLFs is that they must be selected and tuned
\textit{a priori}, without knowledge of how the residuals are actually
distributed.  To address this, \cite{Barron2019} developed an \textit{adaptive}
loss function that adjusts to the shape of the residual distribution,
eliminating the need for model selection and tuning.  This method was recently
extended in \cite{Chebrolu2021}, which made the RLF capable of adapting to a
wider range of distributions.

However, all existing RLFs assume a Gaussian-like distribution with a mode of
zero, and weight highly residuals with a low magnitude.  This is visualized in
\Cref{fig:showcase} by the family of weighting functions that peak at zero.
However, residuals are often defined as the \textit{norm} of a multivariate
error, for example the Mahalanobis distance.  If the error follows a
Gaussian-like distribution, then it's norm will follow a Chi-like distribution
with a mode value $\tilde{\epsilon}$ greater than zero.  This leads to the
development of a ``mode gap,'' shown in \Cref{fig:showcase}, which can
inadvertently influence the weighting scheme.

In \Cref{fig:showcase}, residuals \textit{less} than the mode value are clearly
inliers, as they are expected to be observed given the statistics on error
$\mbf{e}_i$.  When they occur, outliers will produce residual values far greater
than $\tilde{\epsilon}$, in the red region of \Cref{fig:showcase}.  However, due
to the mode gap, \textit{both inliers and outliers receive a moderate to low
weight from RLFs that assume a mode of zero}.  Ambiguous weighting between
inliers and outliers will lead to slower convergence behaviour and less accurate
results, especially if the mode gap grows larger or if the selected RLF is more
conservative, for example Welsch loss. 

The method proposed here, called ``Adaptive MB,'' accounts for this gap by first
estimating the mode value by fitting an $n$-dimensional Maxwell-Boltzmann (MB)
distribution to the residuals.  An MB distribution is a generalization of the
Chi distribution, with a shape parameter that allows for flexibility during the
fitting process.  Accounting for the mode gap, a weight of ${w_i = 1}$ is
assigned to residuals \textit{less} than the mode, which are assumed to be
inliers, and the adaptive method from \cite{Chebrolu2021} is applied to
residuals \textit{greater} than the mode, where outliers actually reside.
\update{The primary contribution of this work is in acknowledging and responding
to this gap, and represents a novel insight in the study of outlier rejection
\cite{DeMenezes2021}.}

The proposed method is tested against several fixed RLFs, as well as existing
adaptive approaches \cite{Barron2019,Chebrolu2021}, on two fundamental state
estimation problems, point cloud alignment and pose averaging.  As predicted
from the theory, accounting for the ``mode gap'' leads to more robust
performance and faster convergence times in these applications.  


\section{Preliminaries}
\label{sec:preliminaries}

\subsection{Robust Loss Functions}
\label{sec:outlier_rejection}

Optimization problems in state estimation often seek to minimize a sum of
\thirdupdate{weighted squared residuals}, 
\vspace{9pt}
\begin{equation}
    \mbf{x}^\star = \argmin_{\mbf{x} \in \rnums^N} J(\mbf{x}) = \argmin_{\mbf{x} \in \rnums^N}  \thirdupdate{\frac{1}{2}} \sum^{N}_{i=1} w_i \cdot \thirdupdate{\epsilon_i} \left( z_i, g_i(x_i) \right)\thirdupdate{^2},
    \label{eqn:weightedopt}%
    \vspace{9pt}
\end{equation}
where ${\mbf{x} \in \rnums^N}$ is the state vector, ${w_i \in
\thirdupdate{(}0,1]}$ are weights, and \thirdupdate{$\epsilon_i$ are residuals}.
The \thirdupdate{residuals} are typically a function of sensor measurements
$z_i$ and some nonlinear function $g_i(x_i)$ of the current state estimate.
\Cref{eqn:weightedopt} is often solved within an \secondupdate{iteratively
reweighted least-squares (IRLS) framework \cite{chartrand2008}, where the
weights $w_i(\thirdupdate{\epsilon_i}(\bar{x}_i))$ are evaluated at the current
operating point $\mbfbar{x}$.  The objective function $J(\mbf{x})$ is then
minimized by linearizing about $\mbfbar{x}$} and applying, for example,
\thirdupdate{Gauss-Newton} or \thirdupdate{Levenberg-Marquardt}
\cite[\textsection4.3.1]{Barfoot2017}.

Robust loss functions modify the objective function in a way that downweights
\thirdupdate{residuals} caused by outliers.  Incorporating robust loss
$\rho(\thirdupdate{\epsilon})$ into \eqref{eqn:weightedopt} produces a new
objective function, 
\vspace{6pt}
\begin{equation}
    J_{\textrm{mod.}}(\mbf{x}) = \sum^N_{i=1} \rho_i(\thirdupdate{\epsilon_i}(x_i)).
    \label{eqn:jmod}%
    \vspace{6pt}
\end{equation}
Rather than minimize \eqref{eqn:jmod} directly, the weights $w_i$ in the
original problem are obtained by equating the gradients, 
\vspace{6pt}
\begin{subequations}
    \begin{align}
        \pd{}{x_i} J(\mbf{x})                 =& \ w_i  \thirdupdate{\cdot} \thirdupdate{\epsilon_i}(x_i) \pd{\thirdupdate{\epsilon_i}(x_i)}{x_i}, \\[6pt]
        \pd{}{x_i} J_{\textrm{mod.}}(\mbf{x}) =& \ \pd{\rho_i(\thirdupdate{\epsilon_i}(x_i))}{\thirdupdate{\epsilon_i}(x_i)} \pd{\thirdupdate{\epsilon_i}(x_i)}{x_i}, \\[-9pt]
        \nonumber
    \end{align}
\end{subequations}
and therefore
\vspace{6pt}
\begin{equation}
    w_i = \frac{1}{\thirdupdate{\epsilon_i}(x_i)} \pd{\rho_i(\thirdupdate{\epsilon_i}(x_i))}{\thirdupdate{\epsilon_i}(x_i)}.
    \vspace{6pt}
\end{equation}
%

\subsection{An Adaptive Robust Loss Function}
\label{sec:adaptivelossbarron}

Recently, \cite{Barron2019} developed an \textit{adaptive} robust loss function,
which dynamically adjusts to match the actual residual distribution.  Defining
the unitless \textit{residual} value as the error scaled by a noise bound $c$, 
\begin{equation}
    \epsilon_i = \frac{e_i}{c},
    \vspace{3pt}
\end{equation}
and accounting for singularities, the RLF is defined piecewise,
\vspace{3pt}
\vspace{-6pt}
\begin{equation}
	\rho(\epsilon, \alpha) =
	\begin{cases}
		\tfrac{1}{2} \epsilon^2 &\text{if} \ \alpha = 2, \\
		\log \left( \tfrac{1}{2} \epsilon^2 + 1 \right) &\text{if} \ \alpha = 0, \\
		1 - \exp \left( -\tfrac{1}{2} \epsilon^2 \right) &\text{if} \ \alpha = -\infty, \\
		\tfrac{|\alpha - 2|}{\alpha} \left( \left( \frac{ \epsilon^2 }{| \alpha - 2 |} + 1 \right)^{\alpha / 2} - 1 \right) &\text{otherwise},
	\end{cases}
	\label{eqn:barronrho}
\end{equation}
where ${\alpha \in (-\infty, 2]}$ is a shape parameter.  To adapt
\eqref{eqn:barronrho}, \cite{Barron2019} assumes the following distribution on
the residuals,
\vspace{6pt}
\begin{equation}
    p(\epsilon \, | \, \mu, \alpha) = \frac{1}{Z(\alpha)} \exp \left( -\rho\left( \epsilon - \mu, \alpha \right) \right),
    \label{eqn:barronp}%
    \vspace{6pt}
\end{equation}
where $\mu$ is the residual mean and $Z(\alpha)$ is a normalization constant.
With ${\mu = 0}$, 
\vspace{6pt}
\begin{equation}
    Z(\alpha) = \! \! \int^\infty_{-\infty} \exp \left( -\rho(\epsilon, \alpha) \right) \, \dee \epsilon.
    \label{eqn:normalizebaron}%
    \vspace{6pt}
\end{equation}
The distribution \eqref{eqn:barronp} is therefore the likelihood of observing
the residuals given shape parameter $\alpha$.  The optimal $\alpha^\star$ is
found by maximizing this likelihood, or alternatively by minimizing the negative
log-likelihood.  For a discrete set of residuals ${\{ \epsilon_i \}^N_{i=1}}$,
\vspace{-9pt}
\begin{subequations}
    \begin{align}
        \alpha^\star =& \, \argmin_{\alpha \in [0, 2]} -\log \left( p(\epsilon \, | \, \alpha) \right) \\
        =& \, \argmin_{\alpha \in [0, 2]} \Big( \underbrace{N \cdot \log( Z(\alpha) ) + \sum^{N}_{i=1} \rho_i(\epsilon_i, \alpha)}_{\Lambda(\alpha)} \Big).
        \label{eqn:findalphastar}%
    \end{align}
    \label{eqn:minlogp}%
\end{subequations}
Having optimized the shape parameter, the weights ${w_i \in (0, 1]}$ in the
original optimization problem \eqref{eqn:weightedopt} are given by
\vspace{3pt}
\begin{equation}
	w_i (\epsilon_i, \alpha^\star) =
	\begin{cases}
		1 &\text{if} \ \alpha^\star = 2, \\
		\tfrac{1}{\sfrac{\epsilon^2_i}{2} + 1} &\text{if} \ \alpha^\star = 0, \\
		\exp \left( -\tfrac{1}{2} \epsilon^2_i \right) &\text{if} \ \alpha^\star = -\infty, \\
		\left( \frac{\epsilon_i^2}{|\alpha^\star-2|} + 1 \right)^{\sfrac{\alpha^\star}{2} - 1} &\text{otherwise}.
	\end{cases}
	\label{eqn:theweights}%
    \vspace{3pt}
\end{equation}
More recently, \cite{Chebrolu2021} noted that the optimization for
$\alpha^\star$ in the original approach is limited to ${\alpha^\star \in
[0,2]}$, as the integral in \eqref{eqn:normalizebaron} is unbounded for ${\alpha
< 0}$.  To address this, \cite{Chebrolu2021} truncates the distribution such
that $-\tau \le \epsilon \le \tau$, 
\vspace{3pt}
\begin{subequations}
    \begin{align}
        \tilde{p}(\epsilon \, | \, \alpha) =& \ \frac{1}{\tilde{Z}(\alpha)} \exp \left( -\rho(\epsilon, \alpha) \right), \\[3pt]
        \vspace{3pt}
        \tilde{Z}(\alpha) =& \! \int^\tau_{-\tau} \exp \left( -\rho(\epsilon, \alpha) \right) \, \dee \epsilon.
        \label{eqn:chebint} \\[-15pt]
        \nonumber
    \end{align}
    \label{eqn:chebrolup}%
\end{subequations}
\secondupdate{In this work, the truncation bound $\tau$ is set according to the
specifics of the problem, and is increased both when the magnitude of the
residuals is expected to be large and when the dimension of the error
increases.}  Using $\tilde{Z}(\alpha)$ in place of $Z(\alpha)$ in
\eqref{eqn:findalphastar}, the optimal shape parameter $\alpha^\star$ is found
here using Newton's method with a backtracking line search, with 
\begin{equation}
    \frac{\dee}{\dee \alpha} \Lambda(\alpha) = -\frac{N}{\tilde{Z}(\alpha)} \int^\tau_{-\tau} \exp(-\rho(\epsilon, \alpha)) \pd{\rho}{\alpha} \dee \epsilon + \sum^N_{i=1} \pd{\rho_i}{\alpha}.
    \label{eqn:deelambdadeealpha}
\end{equation}
This was found to halve the execution time over the grid search method
originally used in \cite{Chebrolu2021}.

\section{Methodology}
\label{sec:methodology}

\subsection{Defining Residuals in Multivariate Least-Squares Problems}
\label{sec:leastsquaresresidual}

More often, state estimation problems seek to minimize a sum of
\secondupdate{weighted, squared \textit{multivariate} errors,}
\vspace{3pt}
\begin{equation}
    J(\mbf{X}) = \frac{1}{2} \sum^N_{i=1} \secondupdate{w_i \cdot} \left\Vert \mbf{e}_i \left( \mbf{z}_i, \mbf{g}_i(\mbf{x}_i) \right) \right\Vert^2_{\mbs{\Sigma}^{-1}_i},
    \label{eqn:multivariatej}%
    \vspace{3pt}
\end{equation}
where ${\mbf{X} = \{ \mbf{x}_i \}^N_{i=1}}$ is a set of states ${\mbf{x}_i \in
\rnums^{n_x}}$, and ${\epsilon^2_i = \| \mbf{e}_i \|^2_{\mbs{\Sigma}\inv_i}}$ is
the squared Mahalanobis distance,
\begin{equation}
    \epsilon^2_i = \mbf{e}_i^\trans \mbs{\Sigma}\inv_i \mbf{e}_i \in \rnums_{\geq 0}.
    \label{eqn:mdistsquared}%
    \vspace{4pt}
\end{equation}
Multivariate error ${\mbf{e}_i \in \rnums^{n_e}}$ is usually assumed to follow a
zero-mean Gaussian distribution, ${\mbf{e}_i \sim \mathcal{N}(\mbf{0},
\mbs{\Sigma}_i)}$, and may contain terms with different units.  In contrast, the
Mahalanobis distance scales each component of $\mbf{e}_i$ according to the
inverse covariance, producing a scalar, unitless residual that accounts for
cross-covariance terms.  Rejecting measurements on the basis of a large
Mahalanobis distance is therefore a theoretically sound way to perform outlier
rejection \cite{Neira2001}.

Taking the Mahalanobis distance as the residual presents a problem for existing
RLFs, for when the components of $\mbf{e}_i$ are assumed to be normally
distributed about zero, $\epsilon_i$ will follow a \textit{Chi distribution}
with mode value ${\tilde{\epsilon} = \sqrt{n_e - 1}}$
\cite[\textsection11.3]{Forbes2010}.  The residuals from inlier measurements
will cluster around $\tilde{\epsilon}$, far from the highly-weighted region
around zero, producing the ``mode gap'' shown in \Cref{fig:showcase}.  As a
result, residuals that are \textit{expected to be observed} given the statistics
on $\mbf{e}_i$ will receive a low weight from existing RLFs.  At best, this gap
will lead to slower convergence times, provided inliers still receive a
relatively high weight compared to outliers.  At worst, inliers and outliers
will receive approximately the same (very low) weight, at which point the
accuracy of the optimization will simply depend on the relative number of
inliers and outliers.  In all cases, these problems are expected to become more
pronounced as the dimension of the error $n_e$, and thus the size of the mode
gap, increases.

\subsection{Mind the Gap: How to Avoid Downweighting Inliers}
\label{sec:dontdownweightinliers}

The present approach addresses this problem by applying the adaptive weighting
method from \cite{Chebrolu2021} to the \textit{mode-shifted} residuals ${\xi_i =
\epsilon_i - \tilde{\epsilon}}$.  In \Cref{fig:showcase}, these are residuals
\textit{greater} than the mode value.  All residuals \textit{less} than the mode
value are treated as \textit{inliers}, and assigned a weight of ${w_i = 1}$.
This is different from taking ${\mu>0}$ in the original formulation
\eqref{eqn:barronp}, as the residuals are not symmetrically distributed about
the mode value and outliers only occur on one side of the distribution. 

A robust estimate of $\tilde{\epsilon}$ is made by fitting an $n_e$-dimensional
\textit{Maxwell-Boltzmann} (MB) speed distribution to the residuals
\cite[\textsection15.2]{Laurendeau2005},
\begin{equation}
	p_{\textrm{MB}}(\epsilon \, | \, a, n_e) = \frac{1}{a^{n_e} 2^{(\sfrac{n_e}{2}-1)} \Gamma(\sfrac{n_e}{2})} \epsilon^{n_e-1} \exp(-\sfrac{\epsilon^2}{(2a^2)}).
	\label{eqn:ndimmb}%
    \vspace{3pt}
\end{equation}
The MB distribution generalizes the Chi distribution with shape parameter $a$,
and the additional flexibility is useful here because, in reality, $\mbf{e}_i$
is merely expected to be Gaussian-like.

Normally, the optimal shape parameter $a^\star$ would be found by minimizing the
negative log-likelihood of $p_{\textrm{MB}}$ in \eqref{eqn:ndimmb} with respect
to $a$.  However, in practice the presence of outliers leads to a poor fit.  The
optimal shape parameter is found here via 
\vspace{6pt}
\begin{subequations}
    \begin{align}
        a^\star =& \, \argmin_{a \in \rnums_{>0}} \expectq{ q(\epsilon) \cdot (p_{\textrm{MB}} - q(\epsilon))^2} 
        \label{eqn:minexpect} \\ 
        =& \, \argmin_{a \in \rnums_{>0}} \int^\infty_0 \big( q(\epsilon) \cdot (p_{\textrm{MB}} - q(\epsilon) ) \big)^2 \, \dee \epsilon, \\[-9pt]
        \nonumber
    \end{align}
    \label{eqn:findastar}%
\end{subequations}
where $q(\epsilon)$ is the distribution on the residuals.  The form of
\eqref{eqn:minexpect} weights the squared difference between $p_{\textrm{MB}}$
and $q$ by the relative frequency $q$, ensuring a better fit in high-frequency
inlier areas.  If the proportion of outliers is especially high, a threshold
\secondupdate{${\epsilon < \tau}$} is applied before the fitting procedure.
Given a discrete set of residuals ${\{ \epsilon_i \}^N_{i=1}}$, the minimizing
solution is
\begin{equation}
    a^\star = \argmin_{a \in \rnums_{>0}} \underbrace{\sum^K_{k=1} \big( q_k \cdot (p_\textrm{MB} - q_k) \big)^2}_{L(a)},
    \label{eqn:astardiscrete}
\end{equation}
where $k$ are the bins of a normalized histogram.  Parameter $a^\star$ is found
here using Newton's method with a backtracking line search, with 
\begin{equation}
    \frac{\dee}{\dee a} L(a) = 2 \sum^K_{k=1} q^2_k \cdot (p_\textrm{MB} - q_k) \pd{p_{\textrm{MB}}}{a}.
\end{equation}
The estimated mode value is then \cite[\textsection15.2]{Laurendeau2005}
\vspace{3pt}
\begin{equation}
    \tilde{\epsilon} = a^\star \sqrt{n_e-1}.
    \label{eqn:mbmode}%
    \vspace{3pt}
\end{equation}
Applying this shift to residuals greater than the estimated mode as well as to
the truncation bound $\tau$ yields the \textit{mode-shifted residuals} $\xi_i$
and the \textit{mode-shifted truncation bound} $\nu$, 
\begin{subequations}
    \begin{align}
        \xi_i =& \ \epsilon_i - \tilde{\epsilon}, \quad \forall \epsilon_i \ge \tilde{\epsilon}, \\
        \nu =& \ \tau - \tilde{\epsilon}.
    \end{align}
    \label{eqn:shiftresiduals}%
\end{subequations}
The adaptive loss optimization in \eqref{eqn:minlogp} is then performed using
the distribution on the mode-shifted residuals $\tilde{p}(\xi \, | \, \alpha)$,
with the integrals in the modified normalization constant \eqref{eqn:chebint}
and gradient \eqref{eqn:deelambdadeealpha} evaluated between $0$ and $\nu$.
Finally, the weights are
\begin{equation}
	\tilde{w}_i(\epsilon_i, \alpha^\star) = 
	\begin{cases}
		1 &\text{if} \ \epsilon_i < \tilde{\epsilon}, \\
		w_i(\xi_i, \alpha^\star) & \text{otherwise},
	\end{cases}
    \label{eqn:mbweights}%
    \vspace{6pt}
\end{equation}
where ${w_i \in (0,1]}$ are found using \eqref{eqn:theweights}.  An example of
this weighting scheme is shown in \Cref{fig:showcase}.

\subsection{Implementing Norm-Aware Adaptive Robust Loss}
\label{sec:summary}

The following steps summarize how to implement norm-aware adaptive robust loss.
\begin{enumerate}
    \item Compute the optimal MB shape parameter $a^\star$ using
    \eqref{eqn:astardiscrete}.
    \item Compute the mode of the MB distribution $\tilde{\epsilon}$ using
    \eqref{eqn:mbmode}.  
    \item Compute the \thirdupdate{mode}-shifted quantities $\xi_i$ and $\nu$ using
    \eqref{eqn:shiftresiduals}.
    \item Compute the optimal adaptive shape parameter $\alpha^\star$,
    \begin{equation}
        \alpha^\star = \argmin_{\alpha \in (-\infty, 2]} M \cdot \log( \tilde{Z}(\alpha) ) + \sum^{M}_{i=1} \rho_i(\xi_i, \alpha),
    \end{equation}
    where ${M < N}$ and where normalization constant is now
    \vspace{3pt}
    \vspace{-10pt}
    \begin{equation}
        \tilde{Z}(\alpha) = \! \! \int^{\nu}_0 \exp \left( -\rho(\xi, \alpha) \right) \, \dee \xi.
        \vspace{3pt}
    \end{equation}
    \item Compute the weights $\tilde{w_i}$ using \eqref{eqn:mbweights} and
    \eqref{eqn:theweights}. 
\end{enumerate}
The weights are then incorporated into the original least-squares problem
\eqref{eqn:multivariatej}, and are evaluated at each step \secondupdate{of an
iteratively reweighted least-squares (IRLS) algorithm.}

\section{Results}
\label{sec:results}

\subsection{Defining Pose Error in the Matrix Lie Algebra}
\label{sec:notation}

Forthcoming results report pose error in the matrix Lie algebra associated with
matrix Lie group $SE(3)$.  This subsection explains how to interpret these
results.   

The position of point $z$ relative to point $w$, resolved in reference frame
$\rframe{a}$, is denoted by ${\mbf{r}^{zw}_a \in \rnums^3}$.  The attitude of
$\rframe{a}$ relative to $\rframe{b}$ is given by direction cosine matrix
$\mbf{C}_{ab}$, with ${\mbf{C} \in SO(3) = \left\{ \mbf{C} \in \rnums^{3\times
3} \, | \, \mbf{C}\mbf{C}^\trans = \eye, \det \mbf{C} = +1 \right\}}$.  Point
$z$ is affixed to a moving object, while point $w$ is stationary in the world.
Frame $\rframe{b}$ rotates with the object, while $\rframe{a}$ remains fixed in
the world.  

An object's position and attitude, collectively called ``pose,'' may be
represented as an element of matrix Lie group $SE(3)$, 
\vspace{3pt}
\vspace{-10pt}
\begin{equation}
    \mbf{T}^{zw}_{ab}(\mbf{C}_{ab}, \mbf{r}^{zw}_a) = \begin{bmatrix}
        \mbf{C}_{ab} & \mbf{r}^{zw}_a \\ \mbf{0} & 1
    \end{bmatrix} \in SE(3),
    \vspace{3pt}
\end{equation}
with ${SE(3) = \left\{ \mbf{T} \in \rnums^{4\times 4} \, | \, \mbf{C} \in SO(3),
\mbf{r} \in \rnums^3 \right\}}$ \cite[\textsection7.1.1]{Barfoot2017}.  Errors
on $SE(3)$ are represented in the matrix Lie algebra ${\mathfrak{se}(3)
\triangleq T_\eye SE(3)}$, defined as the tangent space at the group identity
\cite{Sola2018}.  An element of $\mathfrak{se}(3)$ is given by
\cite[Sec.~2.3]{Arsenault2019}
\vspace{3pt}
\begin{equation}
    \mbs{\xi}^\wedge = \begin{bmatrix}
        \mbs{\phi} \\ \mbs{\rho}
    \end{bmatrix}^\wedge = 
        \begin{bmatrix}
        0 & -\phi_3 & \phi_2 & \rho_1 \\ 
        \phi_3 & 0 & -\phi_1 & \rho_2 \\ 
        -\phi_2 & \phi_1 & 0 & \rho_3 \\ 
        0 & 0 & 0 & 0
    \end{bmatrix} \in \mathfrak{se}(3),
    \vspace{3pt}
\end{equation}
where ${(\cdot)^\wedge : \rnums^6 \to \mathfrak{se}(3)}$ is an isometric
operator.  A matrix Lie group and its corresponding Lie algebra are related
through the matrix exponential and matrix logarithm
\cite[\textsection7.1.3]{Barfoot2017},
\begin{equation}
    \mbf{T} = \exp(\mbs{\xi}^\wedge), \quad \mbs{\xi}^\wedge = \log \left( \mbf{T} \right).
\end{equation}
Attitude and position errors from pose error $\delta \mbf{T}$ are then
\begin{equation}
    \delta \mbs{\xi} = \begin{bmatrix}
        \delta \mbs{\phi}^\trans & \delta \mbs{\rho}^\trans
    \end{bmatrix}^\trans = \log \left( \delta \mbf{T} \right)^\vee \in \rnums^6,
    \label{eqn:lierror}
\end{equation}
where $(\cdot)^\vee$ is the inverse of $(\cdot)^\wedge$, such that
${(\mbs{\xi}^\wedge)^\vee = \mbs{\xi}}$.  Results are reported as the norm of
these errors, $\| \delta \mbs{\phi} \|$ and $\| \delta \mbs{\rho} \|$.

\subsection{Iterative Point Cloud Alignment}
\label{sec:pointcloudalignment}

Point cloud alignment is a fundamental state estimation problem.  Given target
cloud $\mathcal{T} = \{\mbf{r}^{p_jz_1}_{b_1} \}^{M}_{j=1}$ collected at time
$t_1$ and source cloud $\mathcal{S} = \{\mbf{r}^{p_iz_2}_{b_2} \}^{N}_{i=1}$
collected at time $t_2$, the objective is to find transformation
${\mbf{T}^\star_{12} = \bigl( \mbf{T}^{z_2z_1}_{b_1b_2} \bigr)^\star \in SE(3)}$
that best aligns $\mathcal{S}$ to $\mathcal{T}$.  This is done by solving 
\begin{equation}
    \mbf{T}^\star_{12} = \argmin_{\mbf{T} \in SE(3)} \sum^{N}_{i=1} \sum^{M}_{j=1} \frac{b_{ij}}{2} \cdot \secondupdate{w_{ij} \cdot} \left\Vert \mbf{e}_{ij} \left( \mbf{T}_{12}, \mbf{r}^{p_iz_2}_{b_2}, \mbf{r}^{p_jz_1}_{b_1} \right) \right\Vert^2_{\mbs{\Sigma}_{ij}\inv},
    \label{eqn:pcalignment}%
\end{equation}
where ${b_{ij} = \{ 0,1 \}}$ are set by an association solver, ${w_{ij} \in
\thirdupdate{(}0,1]}$ are association weights, $\mbf{e}_{ij}$ are association errors, and
$\mbs{\Sigma}_{ij}$ are error covariances.  \Cref{eqn:pcalignment} is generally
nonconvex, and so iterative point cloud alignment proceeds from initial estimate
$\mbfcheck{T}_{12}$ in search of a local minimum.  When point associations are
made between each point in $\mathcal{S}$ and their nearest neighbour(s) in
$\mathcal{T}$, \eqref{eqn:pcalignment} is referred to as ``iterative closest
point,'' or ICP.  ICP is challenging due to outlier correspondences, often the
result of a low overlap ratio between $\mathcal{S}$ and $\mathcal{T}$.  If the
ICP residual is taken to be the \textit{norm} of the association error, then
accounting for the ``mode gap'' may lead to improved ICP performance.

\begin{figure*}[bt]
	\sbox\subfigbox{%
	  \resizebox{\dimexpr0.98\textwidth-1em}{!}{%
		\includegraphics[height=4cm]{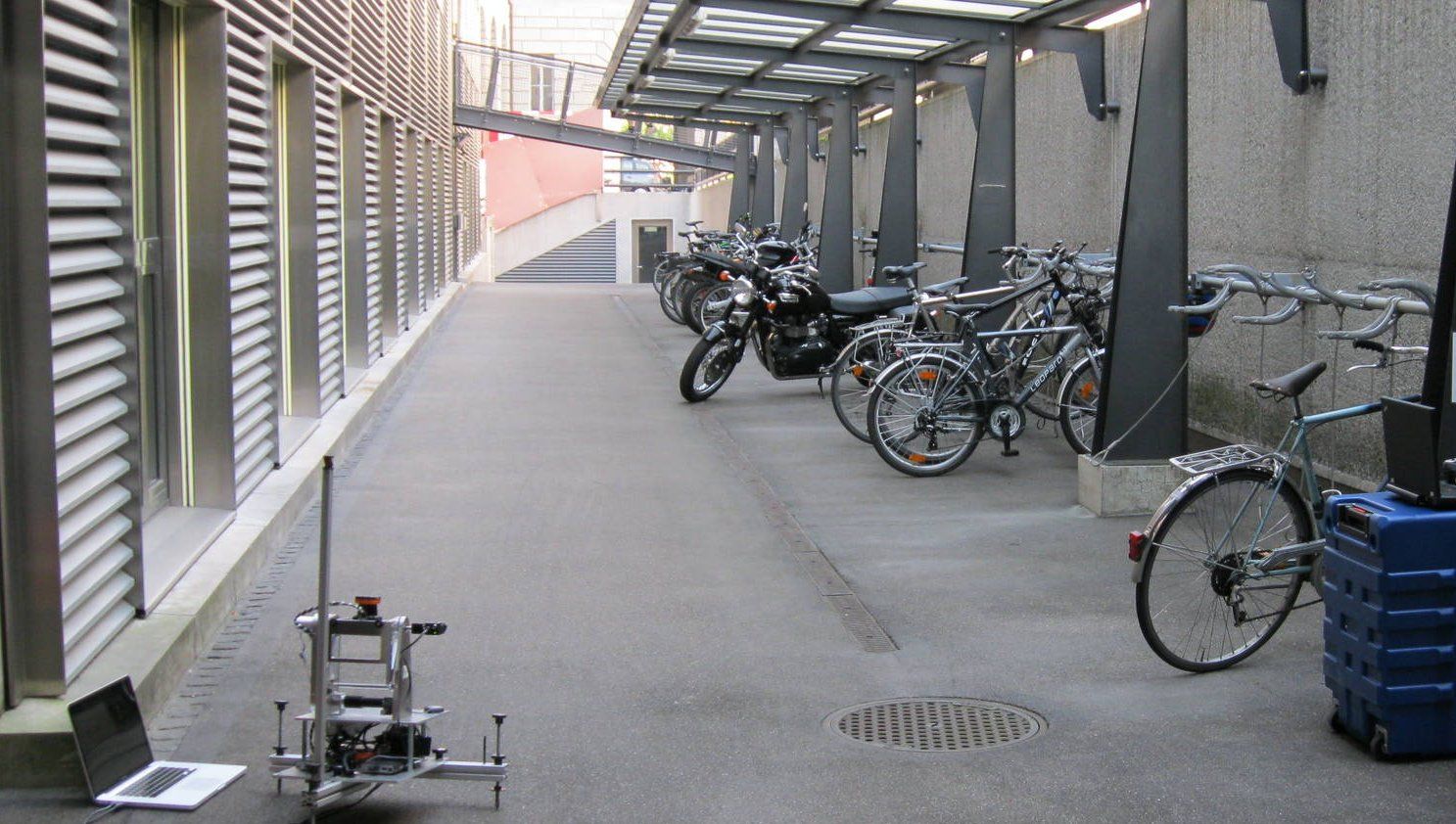}%
		\includegraphics[height=4cm]{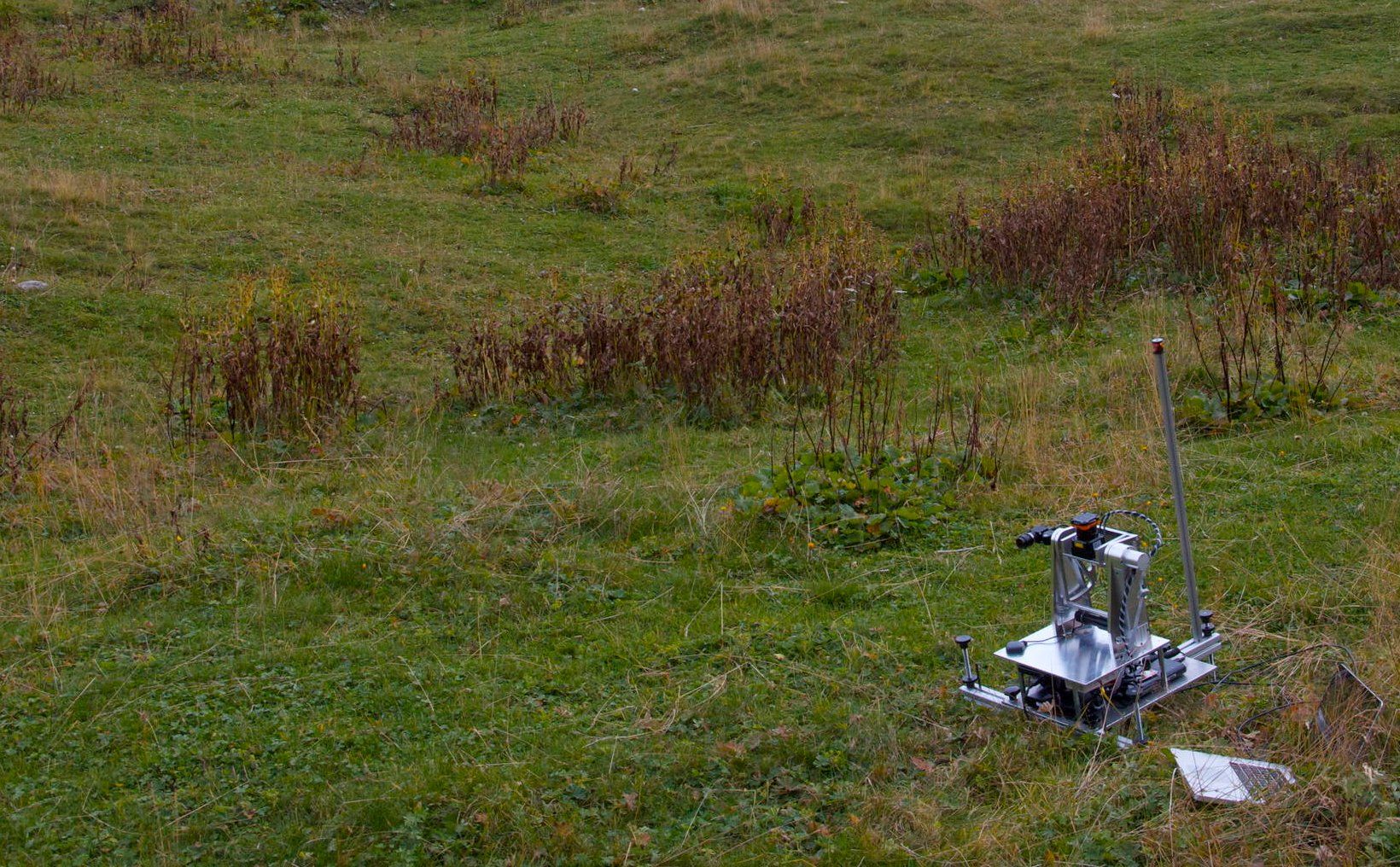}%
		\includegraphics[height=4cm]{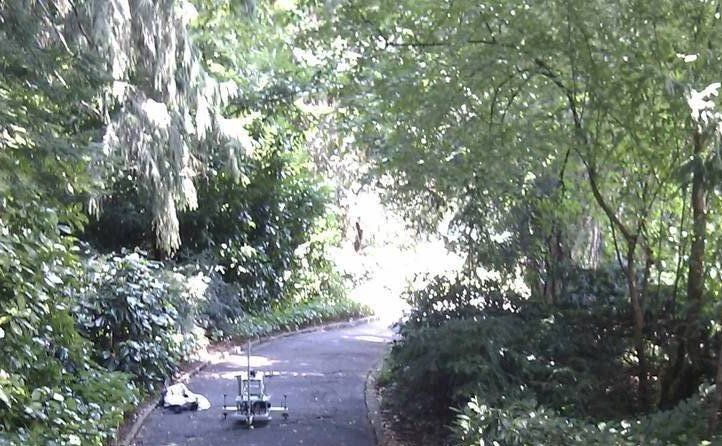}%
	  }%
	}
	\setlength{\subfigheight}{\ht\subfigbox}
	\centering
    \vspace{2pt}    
	\subcaptionbox{Highly structured ``stairs'' (ST) environment    \label{fig:in_st}}{%
	  \includegraphics[height=\subfigheight]{figs/small/env_img_st_5.jpg}
	}
	\subcaptionbox{Semi-structured ``mountain plain'' (MP)  \label{fig:img_mp}}{%
	  \includegraphics[height=\subfigheight]{figs/small/env_img_mp_7.jpg}
	}
	\subcaptionbox{Unstructured ``wood in summer'' (WS)    \label{fig:img_ws}}{%
	  \includegraphics[height=\subfigheight]{figs/small/env_img_ws_5.jpg}
	}%
    \vspace{3pt}
	\subcaptionbox{Two scans from ``stairs''    \label{fig:pc_st}}{%
	  \includegraphics[height=\subfigheight]{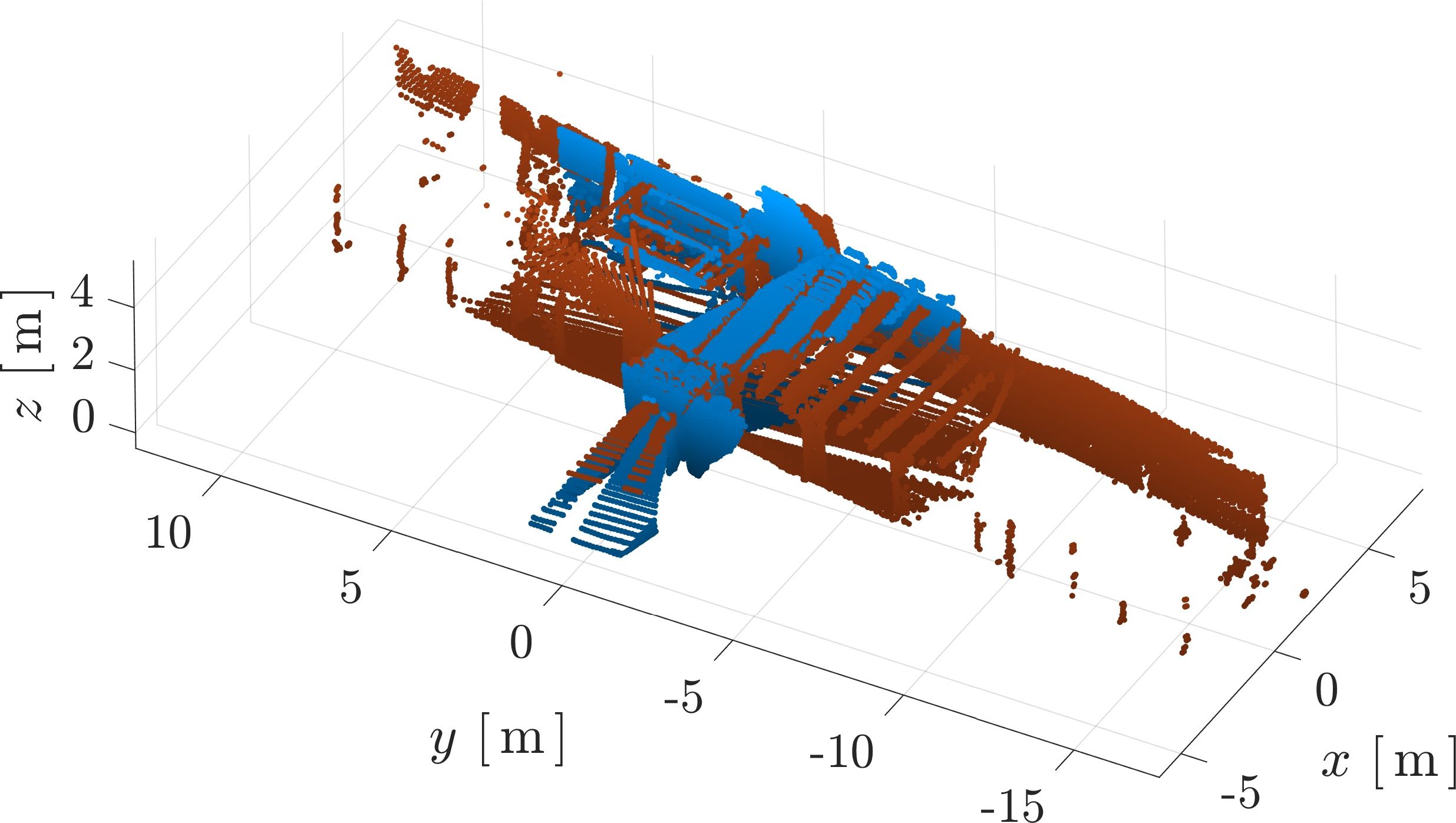}
	}
	\subcaptionbox{Two scans from ``mountain plain''    \label{fig:pc_mp}}{%
	  \includegraphics[height=\subfigheight]{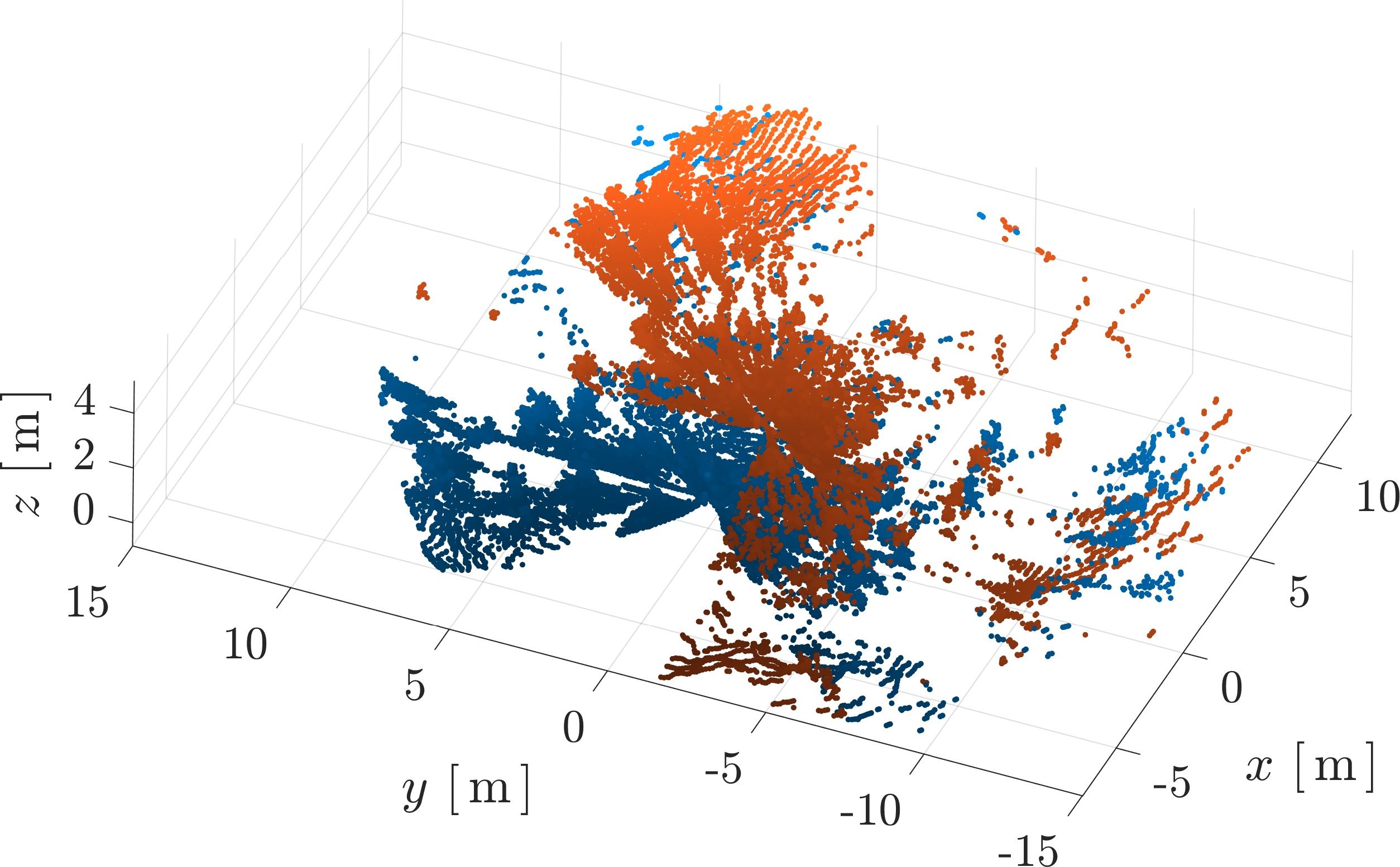}
	}
	\subcaptionbox{Two scans from ``wood in summer''    \label{fig:pc_ws}}{%
	  \includegraphics[height=\subfigheight]{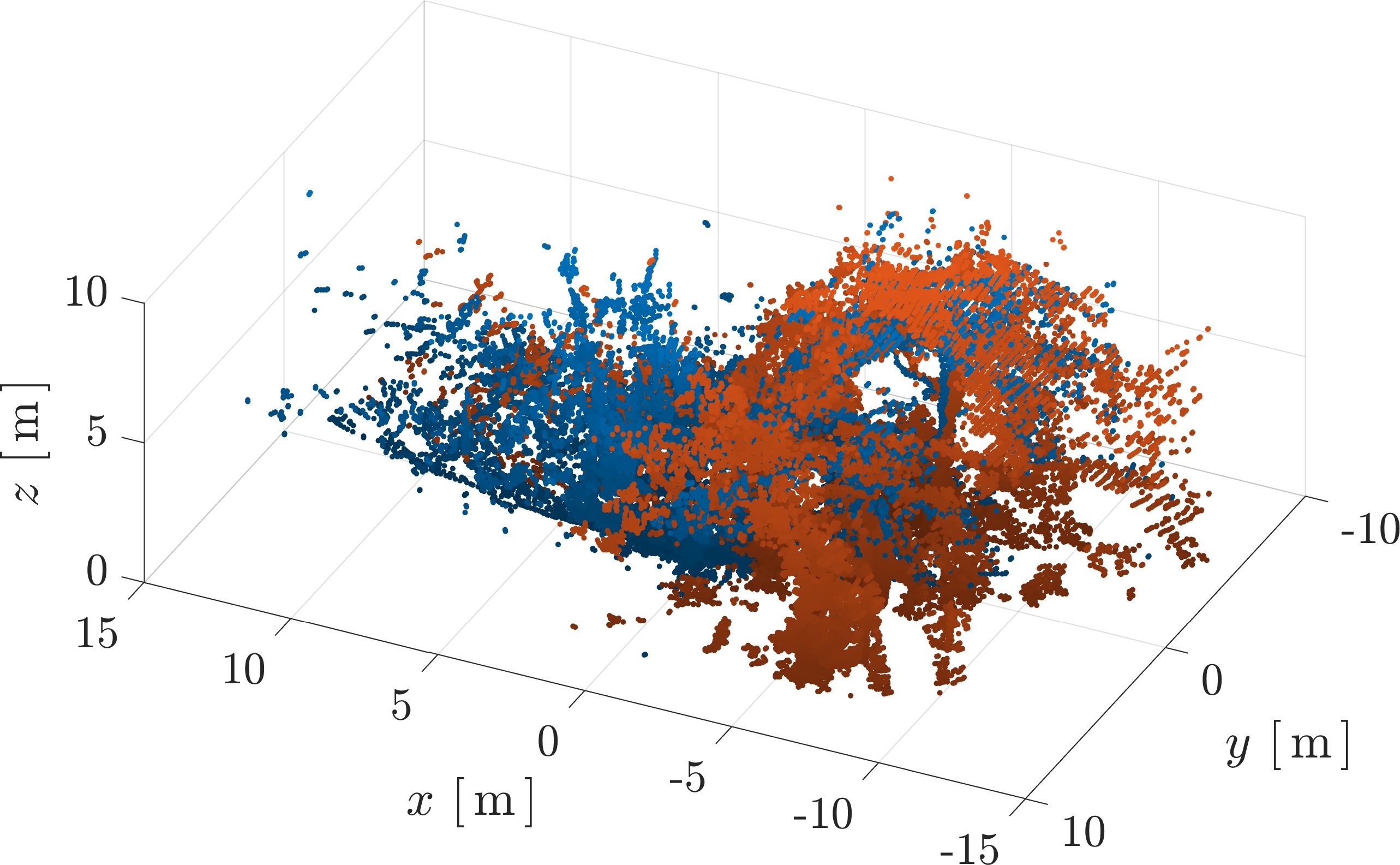}
	}
    \caption{Contextual images and point cloud scans from the different
    environments studied \cite{Pomerleau2012}.  The point cloud scans have been
    downsampled and the orange source cloud perturbed about it's ground-truth
    pose relative to the blue target cloud.  }
    \label{fig:pomerleau_datasets}
\end{figure*}

To investigate this, an ICP experiment was performed on a set of challenging
open-source point cloud datasets \cite{Pomerleau2012}.  Three datasets were
selected: ``stairs'' (ST), ``mountain plain'' (MP), and ``wood in summer'' (WS),
representing structured, semi-structured, and unstructured environments,
respectively.  The datasets contain multiple scans collected from a mobile
platform, as well as ground-truth information and the overlap ratio between scan
pairs.  Contextual images and example scan pairs from the three datasets are
shown in \Cref{fig:pomerleau_datasets}.  

To perform one trial, two scans from an environment are randomly assigned to
$\mathcal{S}$ and $\mathcal{T}$.  The scans are selected across trials to
provide a uniform sampling of the overlap ratio between \SI{40}{\percent} and
\SI{99}{\percent}.  ICP is then initialized according to ${\mbfcheck{T}_{12} =
\mbf{T}_{12} \delta \mbfcheck{T}}$, where $\mbf{T}_{12}$ is the ground-truth
pose of $\mathcal{S}$ relative to $\mathcal{T}$ and $\delta \mbfcheck{T}(\delta
\mbfcheck{C}(\delta \mbscheck{\phi}), \delta \mbfcheck{r})$ is a random
perturbation generated from
\begin{equation}
    \delta \mbscheck{\phi} \sim \mathcal{N} (\mbf{0}, \sigma^2_\phi \eye), \quad \delta \mbfcheck{r} \sim \mathcal{N} (\mbf{0}, \sigma^2_{\textrm{r}} \eye).
\end{equation}
Parameters $\sigma_\phi$ and $\sigma_{\textrm{r}}$ are set such that ${\| \delta
\mbscheck{\phi} \| < \phi_{\textrm{max.}}}$ and ${\| \delta \mbfcheck{r} \| <
r_{\textrm{max.}}}$ for \SI{99.73}{\percent} of the trials.  These limits are
set to ${\phi_{\textrm{max.}} = \SI{20}{\deg}}$ and ${r_{\textrm{max.}} =
\SI{0.5}{\meter}}$, generally corresponding to a ``medium'' level of
perturbation difficulty \cite{Pomerleau2013}.  \secondupdate{To accomodate the
large residuals initially expected as the result of a poor prior relative pose
estimate, the truncation bound is set to ${\tau = 40}$ for all adaptive
methods.}

$\mathcal{S}$ is then aligned to $\mathcal{T}$ within an ICP framework using
different RLFs.  ICP settings are given in \Cref{tab:pcparams}, \update{with
subsampling performed in the sensor frame}.  Seven RLFs are included in this
study.  Cauchy \cite{Black1996}, Tukey \cite{Beaton1974}, and Welsch
\cite{Dennis1978} loss functions implemented with median of absolute deviation
(MAD) rescaling \cite{Haralick1989}, as well as the Var. Trimmed loss function
\cite{Phillips2007}, are fixed functions included for their good performance in
a similar study \cite{Babin2019}.  ``Adaptive Barron'' is the original adaptive
function \cite{Barron2019}, and ``Adaptive Chebrolu'' is the recent modification
\cite{Chebrolu2021}.  These are compared to the proposed approach, called
``Adaptive Maxwell-Boltzmann (MB).''

While a squared point-to-plane error is minimized by ICP, the residual
${\epsilon_i = \small \sqrt{\partsmash[b]{0.5}{\mbf{e}_{ij}^\trans
\mbs{\Sigma}_{ij}\inv \mbf{e}_{ij}}}}$ used by all RLFs to perform outlier
rejection is constructed using the point-to-point error,
\vspace{3pt}
\begin{subequations}
    \begin{align}
        \mbf{e}_{ij} =& \ \mbf{r}^{p_jz_1}_{b_1} - \left( \mbf{r}^{z_2z_1}_{b_1} + \mbf{C}_{b_1b_2} \mbf{r}^{p_iz_2}_{b_2} \right), \\[3pt]
        \mbs{\Sigma}_{ij} =& \ \mbf{C}_{b_1b_2} \mbf{R}_i \mbf{C}^\trans_{b_1b_2} + \mbf{R}_j,
        \\[-12pt] \nonumber
    \end{align}
\end{subequations}
where ${\mbf{R}_i, \mbf{R}_j}$ are the covariances on point measurements
$\mbf{r}^{p_iz_2}_{b_2}$ and $\mbf{r}^{p_jz_1}_{b_1}$, respectively.
\secondupdate{To reduce the number of residual terms, $\mathcal{S}$ is
subsampled on a \SI{10}{\centi\meter} voxel grid.  The covariance on the point
measurements is set to ${\mbf{R}_i = \mbf{R}_j = \sigma_\ell^2 \eye}$, with
${\sigma_\ell = \SI{3}{\centi\meter}}$, as reported in \cite{Pomerleau2012}.}
Posterior errors $\| \delta \mbshat{\phi} \|$ and $\| \delta \mbshat{\rho} \|$
corresponding to ICP posterior estimate $\mbfhat{T}_{12}$ are reported according
to \eqref{eqn:lierror}, with ${\delta \mbfhat{T} = \mbf{T}_{12}\inv
\mbfhat{T}_{12}}$.  180 trials per RLF were performed in each environment, with
results in \Cref{tab:pomerleau_table_error}.
%
%
\setlength{\tabcolsep}{4pt}
\begin{table}[tb]
    \centering
    \caption{Point cloud preprocessing and alignment parameters}
    \label{tab:pcparams}
    \renewcommand{\arraystretch}{1.2}
    \begin{tabularx}{\columnwidth}{llX}
    \toprule
    \secondupdate{Stage} & Configuration & Description \\ \hline
    Preprocessing & \texttt{VoxelGrid} & Subsample $\mathcal{S}$ on \SI{10}{\centi\meter} grid \\
    & \texttt{Normals} & From 15 nearest neighbours \\
    ICP data assn. & \texttt{KDTree} & Single nearest neighbour \\
    ICP error min. & \texttt{Pt-pl} & Point-to-plane error \\
    Termination & \texttt{Diff.} & ${\| \delta \mbs{\phi}_i \| <
    \SI{1e-3}{\radian} \textrm{, and}}$ ${\| \delta \mbs{\rho}_i \| <
    \SI{1e-3}{\meter}}$ \\
    & \texttt{Counter} & 50 iterations max \\
    \bottomrule
    \end{tabularx}
\end{table}

\begin{figure*}[h!]
    \vspace{36pt}
	\sbox\subfigbox{%
	  \resizebox{\dimexpr0.975\textwidth-1em}{!}{%
		\includegraphics[height=4cm]{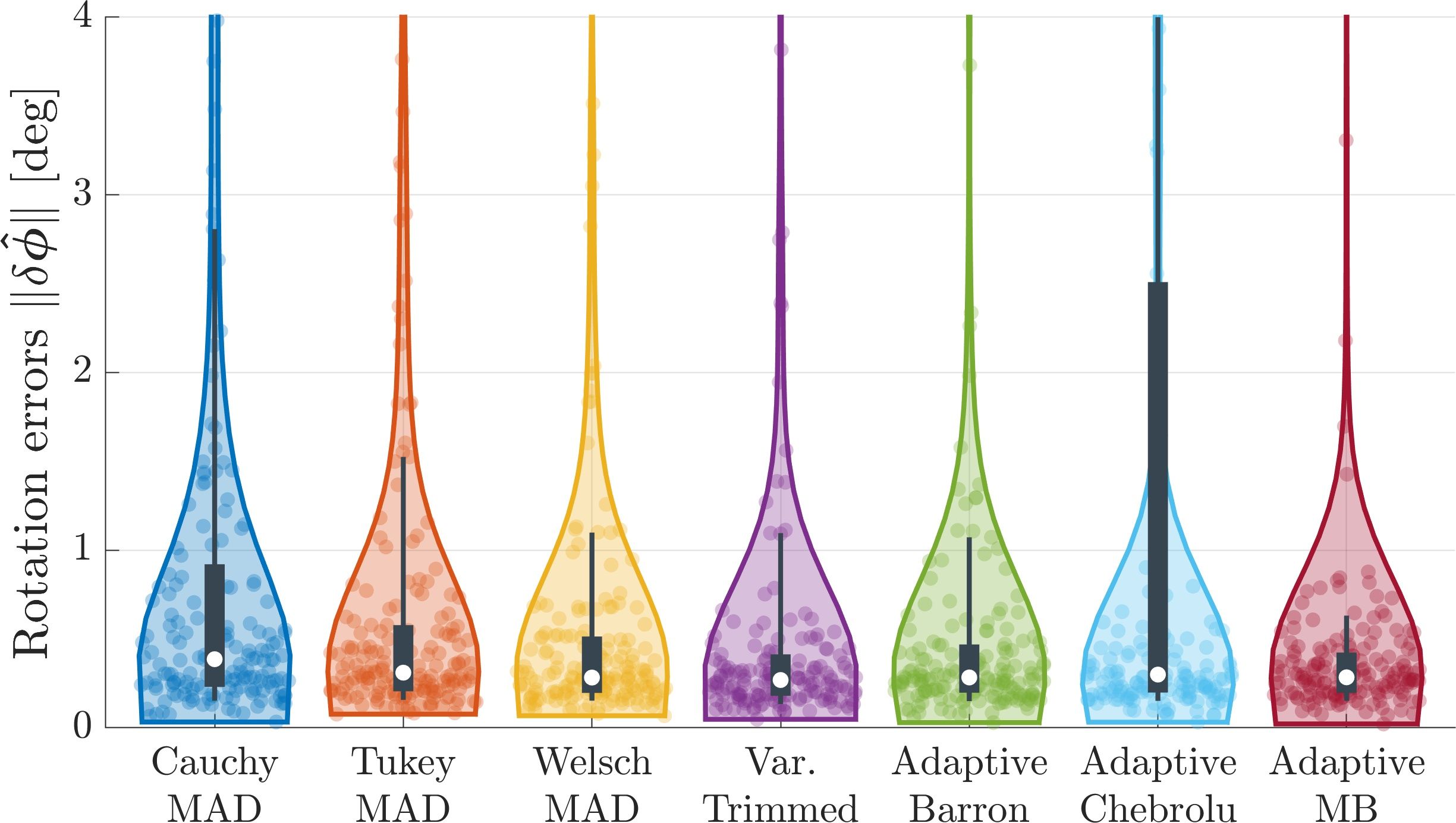}%
		\includegraphics[height=4cm]{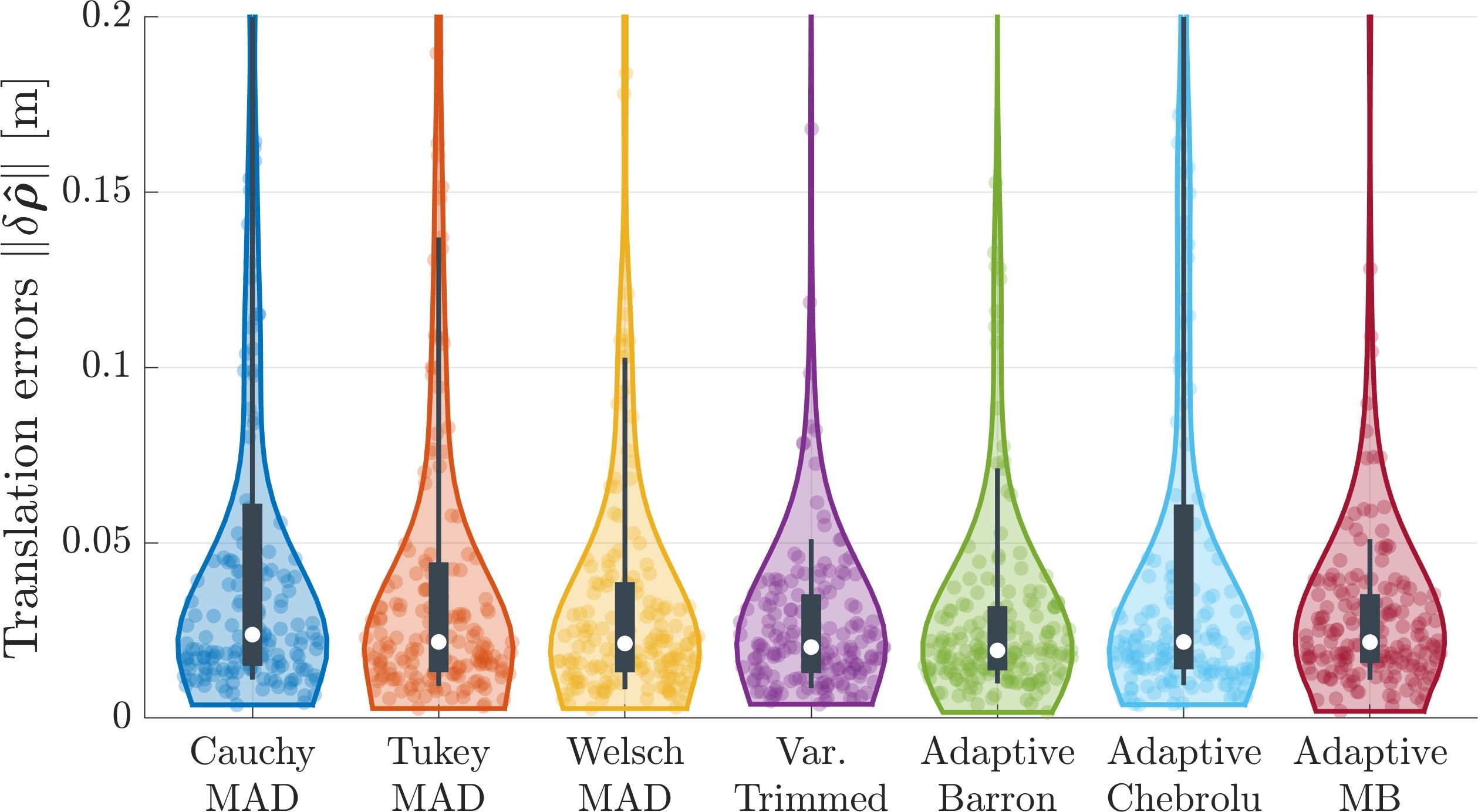}%
	  }%
	}
	\setlength{\subfigheight}{\ht\subfigbox}
	\centering
    \vspace{2pt}    
	\subcaptionbox{Rotation errors, wood in summer (WS) dataset (180 trials) \label{fig:pom_ws_phi}}{%
	  \includegraphics[height=\subfigheight]{figs/mc_results_new_formulation/small/pomerleau_wood_summer_error_phi.jpg}
	}
    \hspace{3pt}
	\subcaptionbox{Translation errors, WS dataset \label{fig:pom_ws_rho}}{%
	  \includegraphics[height=\subfigheight]{figs/mc_results_new_formulation/small/pomerleau_wood_summer_error_rho.jpg}
	}%
    \vspace{6pt}
    \subcaptionbox{Rotation errors, combined across datasets (ALL, 540 trials) \label{fig:pom_all_phi}}{%
	  \includegraphics[height=\subfigheight]{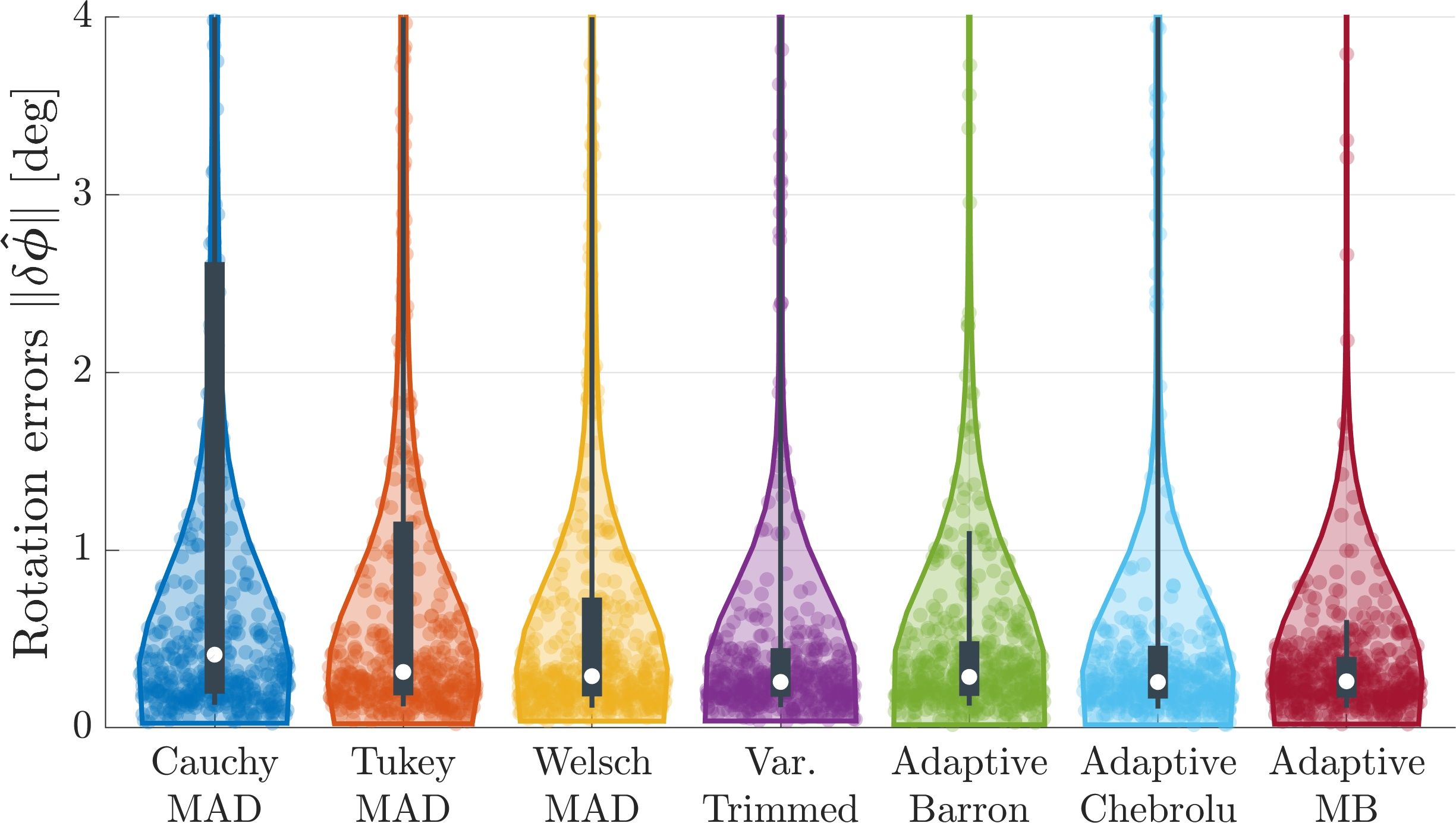}
	}
    \hspace{3pt}
	\subcaptionbox{Translation errors, combined across datasets \label{fig:pom_all_rho}}{%
	  \includegraphics[height=\subfigheight]{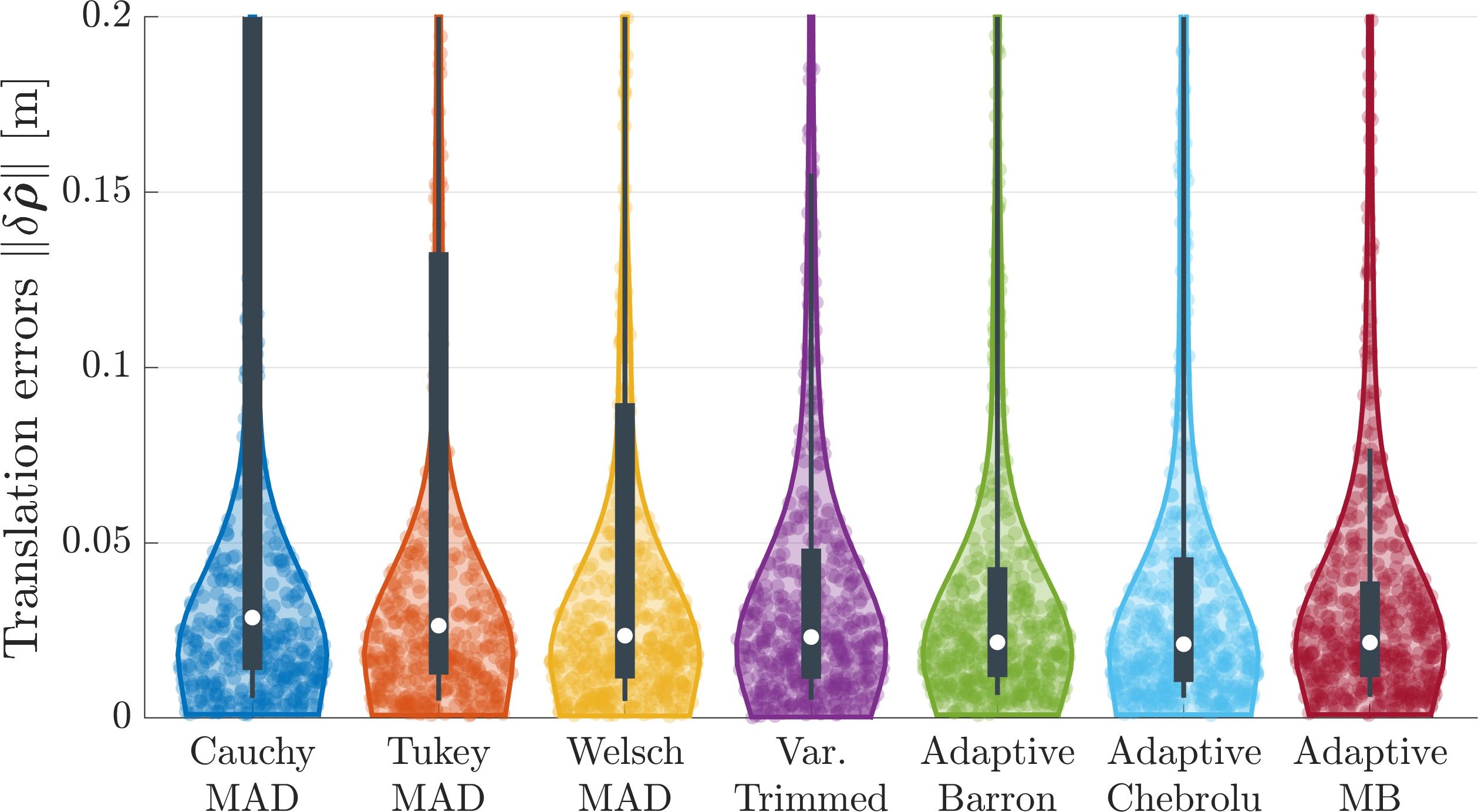}
	}
	\caption{\secondupdate{Violin plots} showing the distribution of ICP alignment errors.
	Scatter plots show errors from individual trials, while envelopes show their
	relative frequency.  Box plots show the middle \SI{50}{\percent} (thick bar)
	and \SI{80}{\percent} (thin bar) of the data, as well as the median (white
	dot).  Median errors for Adaptive MB are slightly lower than
	state-of-the-art adaptive approaches across different environments, however
	the variability is markedly lower, especially for unstructured environments
	like WS (top row).}
    \label{fig:pomerleau_violin}
\end{figure*}

The proposed Adaptive MB approach delivers the lowest median rotation and
translation errors across most experiments.  However, what is especially notable
about this approach is the large reduction in error \textit{variability},
measured by the \SI{75}{\percent} and \SI{90}{\percent} error bounds for the
adaptive RLFs on the right side of \Cref{tab:pomerleau_table_error}.  This
reduction is more pronounced for less structured environments.  For example, on
the highly unstructured ``wood in summer'' (WS) dataset, \SI{90}{\percent} of
translation errors for Adaptive MB are below \secondupdate{\SI{5.2}{\centi
\meter}}, compared with \secondupdate{\SI{7.2}{\centi \meter}} for the original
adaptive RLF \cite{Barron2019} and \secondupdate{\SI{22.8}{\centi \meter}} for
the recent modification \cite{Chebrolu2021}.  The reduction in variability is
also evident for rotation errors, with \SI{90}{\percent} of rotation errors in
the WS dataset falling below \secondupdate{\SI{0.64}{\deg}} for Adaptive MB,
compared with \secondupdate{\SI{1.09}{\deg}} for Adaptive Barron and
\secondupdate{\SI{9.00}{\deg}} for Adaptive Chebrolu.  This reduction in
variability is clearly seen in the violin plots in \Cref{fig:pomerleau_violin}.  

The bottom two rows of \Cref{tab:pomerleau_table_error} give the median number
of iterations to convergence and the median convergence time for all RLFs, taken
across all datasets.  Adaptive MB converges in fewer iterations than
\secondupdate{all the other methods surveyed and, despite the additional
optimization step involved, converges in less time than the other adaptive
methods.}  Note that this study was conducted using non-optimized
\textsc{MATLAB} code, and timing results are included for relative comparison.
On average, the fixed RLFs produced larger \secondupdate{median errors and more
severe failures} than the adaptive RLFs, highlighting the effectiveness of
adaptive methods. 

\Cref{tab:pomerleau_table_performance} shows the success rate of each RLF for
the different datasets.  A trial is considered successful if the RLF is able to
reduce \textit{both} the rotation and translation error over the initial
perturbation.  \secondupdate{With the exception of Var. Trimmed}, the adaptive
RLFs generally outperform the fixed RLFs \secondupdate{across the different
environments, however of all the methods surveyed} the proposed Adaptive MB RLF
has the highest success rate.  

These improvements have been realized solely from a modification to the
association weights $w_{ij}$ in \eqref{eqn:pcalignment}.  Adaptive MB is able to
improve on existing methods because it \textit{avoids downweighting inliers},
leading to lower median errors, lower variability, and fewer iterations to
convergence.

\newcolumntype{Y}{>{\centering\arraybackslash}X}
\newcolumntype{b}{Y}
\newcolumntype{s}{>{\hsize=.5\hsize}Y}
\newcolumntype{P}[1]{>{\centering\arraybackslash}p{#1}}
\begin{table*}[htb]
    \centering
    \vspace{3pt}    
    \caption{ICP alignment errors and timing results for the environments
    studied using different RLFs.  Results are also combined across environments
    (ALL).  Median errors are reported for the four fixed RLFs, while cumulative
    statistics are reported for the adaptive RLFs in the format
    50\%-\colour{dark_grey}{75\%}-\colour{light_grey}{90\%}, corresponding to
    the upper bounds of the error bars shown in \Cref{fig:pomerleau_violin}.
    The lowest median error in each row appears in bold font, as well as the
    lowest 75\% and 90\% errors for the adaptive RLFs.  The median number of
    iterations and execution time are reported across all datasets for relative
    comparison.}
    \renewcommand{\arraystretch}{1.2}
    \begin{tabularx}{\textwidth}{p{1.25cm}P{0.75cm}|ssss|bbb}
    \toprule
    \multicolumn{2}{c|}{\multirow{2}{*}{\secondupdate{Environment}}}  & Cauchy MAD & Tukey MAD & Welsch MAD & Var. Trimmed & Adaptive
    Barron~\cite{Barron2019} & Adaptive Chebrolu~\cite{Chebrolu2021} & Adaptive MB~(ours) \\
    \hline
    \multirow{4}{*}{\begin{tabular}{@{}c@{}} $\| \delta \mbshat{\phi} \|$ \\ $\left[ \deg \right]$ \end{tabular}}
        & ST & 0.29 & 0.27 & 0.24 & 0.23 & 0.23-\colour{dark_grey}{0.41}-\colour{light_grey}{0.83} & \textbf{0.21}-\textbf{\colour{dark_grey}{0.31}}-\colour{light_grey}{0.55} & \textbf{0.21}-\colour{dark_grey}{0.32}-\textbf{\colour{light_grey}{0.48}} \\ 
        & MP & 0.57 & 0.42 & 0.39 & 0.30 & 0.36-\colour{dark_grey}{0.64}-\colour{light_grey}{1.77} & 0.29-\colour{dark_grey}{0.54}-\colour{light_grey}{3.49} & \textbf{0.27}-\textbf{\colour{dark_grey}{0.44}}-\textbf{\colour{light_grey}{0.68}} \\ 
        & WS & 0.39 & 0.31 & 0.28 & \textbf{0.27} & 0.28-\colour{dark_grey}{0.47}-\colour{light_grey}{1.09} & 0.30-\colour{dark_grey}{2.51}-\colour{light_grey}{9.00} & 0.28-\textbf{\colour{dark_grey}{0.42}}-\textbf{\colour{light_grey}{0.64}} \\ 
        & ALL & 0.41 & 0.31 & 0.29 & \textbf{0.26} & 0.29-\colour{dark_grey}{0.48}-\colour{light_grey}{1.11} & \textbf{0.26}-\colour{dark_grey}{0.46}-\colour{light_grey}{5.79} & \textbf{0.26}-\textbf{\colour{dark_grey}{0.40}}-\textbf{\colour{light_grey}{0.61}} \\ 
    \hline
    \multirow{4}{*}{\begin{tabular}{@{}c@{}} $\| \delta \mbshat{\rho} \|$ \\ $\left[ \meter\!\meter \right]$ \end{tabular}}
        & ST & 14 & 12 & 11 & 11 & 11-\colour{dark_grey}{24}-\colour{light_grey}{226} & \textbf{9}-\textbf{\colour{dark_grey}{16}}-\colour{light_grey}{65} & 10-\colour{dark_grey}{18}-\textbf{\colour{light_grey}{34}} \\ 
        & MP & 66 & 47 & 43 & 36 & 36-\colour{dark_grey}{103}-\colour{light_grey}{293} & 35-\colour{dark_grey}{64}-\colour{light_grey}{213} & \textbf{34}-\textbf{\colour{dark_grey}{63}}-\textbf{\colour{light_grey}{151}} \\ 
        & WS & 24 & 22 & 21 & 20 & \textbf{19}-\textbf{\colour{dark_grey}{32}}-\colour{light_grey}{72} & 22-\colour{dark_grey}{61}-\colour{light_grey}{228} & 22-\colour{dark_grey}{35}-\textbf{\colour{light_grey}{52}} \\ 
        & ALL & 29 & 26 & 24 & 23 & 22-\colour{dark_grey}{43}-\colour{light_grey}{217} & \textbf{21}-\colour{dark_grey}{46}-\colour{light_grey}{205} & 22-\textbf{\colour{dark_grey}{39}}-\textbf{\colour{light_grey}{79}} \\ 
    \hline
    Iterations & ALL & 20 & 20 & 22 & 24 & 23 & 26 & 17 \\
    Time $\left[ \SI{}{\second} \right]$ & ALL & 2.75 & 2.75 & 2.94 & 3.38 & 3.64 & 4.21 & 2.83 \\	\bottomrule
    \end{tabularx}
	\label{tab:pomerleau_table_error}
\end{table*}

\begin{table}[tb]
    \centering
    \caption{\secondupdate{Comparing} the percentage of successful alignments
    using different RLFs.  An alignment is considered successful if it reduces
    \textit{both} the rotation and translation error over the initial
    perturbation, ${\| \delta \mbshat{\phi} \| < \| \delta \mbscheck{\phi} \|}$
    and ${\| \delta \mbshat{\rho} \| < \| \delta \mbscheck{\rho} \|}$.}
    \renewcommand{\arraystretch}{1.2}
    \begin{tabularx}{\columnwidth}{p{3.35cm}|YYYY}
    \toprule
        RLF & ST & MP & WS & ALL \\
        \hline
        Cauchy-MAD & 76.7\% & 53.9\% & 85.0\% & 71.9\% \\ 
        Tukey-MAD & 78.9\% & 62.2\% & 91.1\% & 77.4\% \\ 
        Welsch-MAD & 79.4\% & 68.3\% & 93.9\% & 80.6\% \\ 
        Var. Trimmed & 87.8\% & 90.0\% & 96.7\% & 91.5\% \\
        \hline
        Adaptive Barron \cite{Barron2019} & 88.3\% & 85.6\% & 96.1\% & 90.0\% \\ 
        Adaptive Chebrolu \cite{Chebrolu2021} & 92.8\% & 88.9\% & 88.3\% & 90.0\% \\ 
        Adaptive MB (ours) & \textbf{96.1\%} & \textbf{92.2\%} & \textbf{98.3\%} & \textbf{95.6\%} \\ 
        \bottomrule
    \end{tabularx}
	\label{tab:pomerleau_table_performance}
\end{table}

\subsection{Pose Averaging}
\label{sec:poseavg}

Pose averaging is a second fundamental state estimation problem, often
encountered in camera-based applications such as structure from motion (SfM)
\cite{Karimian2020}.  A pose averaging study was performed investigating the
effectiveness of the RLFs in a six degree-of-freedom problem with high outlier
rates. 

Given a set of pose measurements $\{ \mbftilde{T}_i \}\vphantom{\{ \mbf{T}_i
\}}^N_{i=1}$ with associated covariances $\left\{ \mbf{R}
\right\}\!\vphantom{\left\{ \mbf{R} \right\}}^N_{i=1}$, ${\mbf{R}_i =
\expect{\delta \mbstilde{\xi}_i \, \delta \mbstilde{\xi}^\trans_i}}$, pose
averaging returns the optimal average pose, 
\vspace{6pt}
\begin{equation}
    \mbf{T}^\star = \argmin_{\mbf{T} \in SE(3)} \frac{1}{2} \sum^N_{i=1} w_i \cdot \left\Vert \mbf{e}_i(\mbf{T}, \mbftilde{T}_i) \right\Vert^2_{\mbs{\Sigma}\inv_i},
    \label{eqn:poseavgcost}%
    \vspace{6pt}
\end{equation}
with left-invariant pose error ${\mbf{e}_i = \log \left( \mbf{T}\inv \,
\mbftilde{T}_i \right)^\vee}$ \cite[\textsection5.2.1]{Arsenault2019}.
Perturbing to first order yields the batch Jacobians, 
\vspace{6pt}
\begin{equation}
    \delta \mbf{e}_i = \underbrace{\mbf{J}^\ell(\mbfbar{e}_i)\inv}_{\mbf{H}_i} \delta \mbs{\xi}_i \underbrace{- \mbf{J}^{\textrm{r}}(\mbfbar{e}_i)\inv}_{\mbf{M}_i} \delta \mbstilde{\xi}_i,
    \vspace{6pt}
\end{equation}
where $\mbf{J}^\ell, \mbf{J}^\textrm{r}$ are, respectively, the left and right
Jacobians of $SE(3)$ \cite[\textsection7.1.5]{Barfoot2017}.  The covariance on
the pose errors is then ${\mbs{\Sigma}_i = \mbf{M}_i \mbf{R}_i
\mbf{M}^\trans_i}$.  The residual $\epsilon_i$ is defined via
\eqref{eqn:mdistsquared}.  If $\mbf{e}_i$ is normally distributed, then
$\epsilon_i$ will follow a Chi distribution with six degrees of freedom, with a
mode of ${\tilde{\epsilon} = \sqrt{5}}$ \cite[\textsection11.3]{Forbes2010}.  By
accounting for this gap, Adaptive MB is expected to converge faster with better
overall performance than existing RLFs.  

\begin{figure}[t]
	\centering
	\includegraphics[width=\columnwidth]{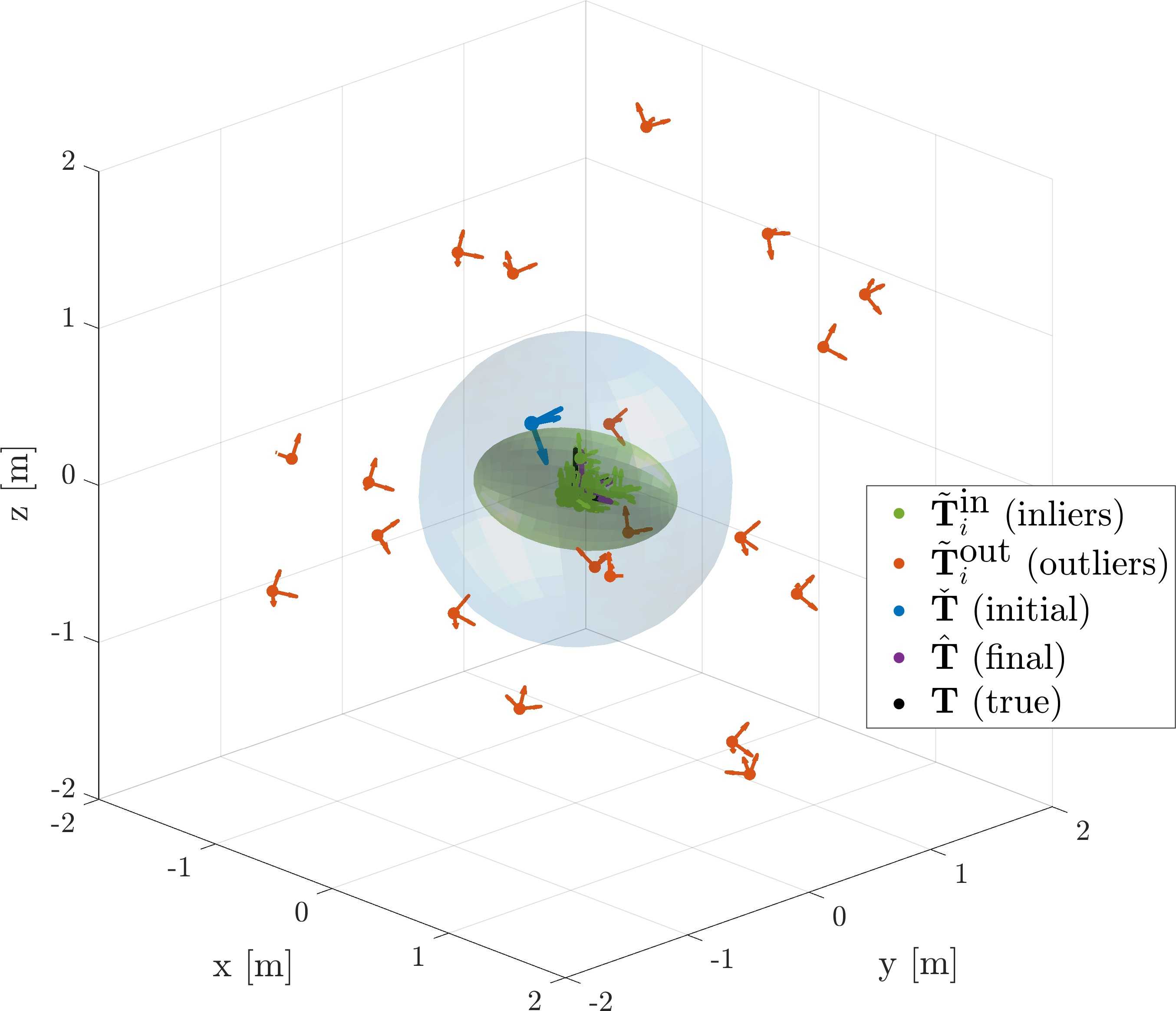}
	\caption{A pose averaging trial with a \SI{60}{\percent} outlier rate.  The
	green and blue ellipsoids are the \SI{99.73}{\percent} confidence bounds on
	$\mbf{R}$ and $\mbfcheck{P}$, respectively.  Note $\mbf{R}$ is
	non-isotropic, highlighting the importance of using the Mahalanobis distance
	as the residual.}
	\label{fig:pose_avg_trial}
    \vspace{-6pt}
\end{figure}

A simulated Monte Carlo experiment was performed in which a pose averaging
problem was corrupted by an increasing number of outliers.  For all trials, 20
inlier measurements were randomly generated according to
${\mbftilde{T}^{\textrm{in}}_i = \exp(\delta \mbstilde{\xi}_i^\wedge)}$, with
${\delta \mbstilde{\xi}_i \sim \mathcal{N}(\mbf{0}, \mbf{R})}$.  Outlier
measurements were generated uniformly according to
${\mbf{T}^\textrm{out}_i(\mbf{C}_i(\mbs{\phi}_i^{\textrm{out}}),
\mbf{r}_i^\textrm{out})}$, where ${\mbs{\phi}^\textrm{out}_i \in \left[
-\SI{60}{\deg}, \SI{60}{\deg} \right]}$ and ${\mbf{r}^\textrm{out}_i \in \left[
-\secondupdate{\SI{2.5}{\meter}}, \secondupdate{\SI{2.5}{\meter}} \right]}$.
The proportion of outlier measurements was increased from \SI{20}{\percent} to
\SI{80}{\percent} in \SI{20}{\percent} increments.  100 trials were run at each
outlier level, with $\mbfcheck{T}$ initialized according to ${\mbfcheck{T} =
\exp(\delta \mbscheck{\xi}^\wedge)}$, ${\delta \mbscheck{\xi} \sim \mathcal{N}
(\mbf{0}, \mbfcheck{P})}$.  A trial converged when the update fell below the
thresholds ${\| \delta \mbs{\phi}_i \| < \SI{1e-3}{\radian}}$ and ${\| \delta
\mbs{\rho}_i \| < \SI{1e-3}{\meter}}$.  Trials were terminated after 50
iterations.  \Cref{fig:pose_avg_trial} shows the setup for a single trial,
including a visualization of $\mbf{R}$ and $\mbfcheck{P}$.  The truncation bound
$\tau$ was set to \secondupdate{40} for all adaptive approaches.  The pose
averaging experiment is intentionally simple, to evaluate the behaviour of the
different RLFs under tightly controlled conditions.

The results are summarized in \Cref{tab:pose_avg_table}.  Adaptive Chebrolu
\cite{Chebrolu2021} performs reasonably well at low outlier levels, but becomes
increasingly conservative ${(\alpha^\star \to -\infty)}$ as the proportion of
outliers is increased, resembling Welsch loss.  The effect on the optimization
is twofold.  First, the (low) weights assigned to the inlier residuals clustered
around the mode at ${\tilde{\epsilon} = \sqrt{5}}$ become increasingly
indistinguishable from the weights assigned to the outlier residuals.  Second,
the number of outlier terms grows relative to the number of inlier terms.  These
effects combine to yield reduced performance at higher outlier levels.

The original Adaptive Barron \cite{Barron2019} approach generally avoids this,
as the shape parameter is bounded from below by zero, resembling Cauchy loss.
The optimization is better able to differentiate inliers from outliers by the
relative magnitude of their association weight, leading to reasonable
performance.  However, the weights assigned to the inliers are still low,
leading to slower convergence.

\secondupdate{In contrast, Adaptive MB acknowledges the ``mode gap,'' assigning
an association weight of ${w_i = 1}$ to inlier residuals ${\epsilon_i <
\tilde{\epsilon}}$.  This approach converges more quickly than the other
adaptive methods, with a median execution time of \SI{0.13}{\second}.  In nearly
all trials, accounting for the underlying distribution on $\epsilon_i$ produces
the lowest median error, with a marked decrease in error variability compared to
other adaptive methods.}

\secondupdate{Notably, Var. Trimmed also yields excellent median performance
across all pose averaging trials, with a much lower median execution time than
Adaptive MB.  This truncated loss function may therefore prove a useful
alternative for time-sensitive applications.}

\begin{table*}[tb]
    \centering
    \vspace{3pt}    
    \caption{Pose errors and timing results from the simulated pose averaging
    experiment, showing either median values or a
    50\%-\colour{dark_grey}{75\%}-\colour{light_grey}{90\%} distribution.  As
    the proportion of outliers goes up, Adaptive Chebrolu \cite{Chebrolu2021}
    grows increasingly conservative ($\alpha^\star \to -\infty$), resembling
    Welsch loss.  However, because the mode of the residuals occurs at
    ${\tilde{\epsilon} > 0}$,  \textit{both inliers and outliers are assigned an
    approximately equal (low) weight}, yielding poorer performance.  Adaptive
    Barron \cite{Barron2019} generally avoids this, as $\alpha$ is bounded from
    below by zero, resembling Cauchy loss.  Adaptive MB converges faster than
    the other adaptive methods, both in terms of iterations and execution time,
    because \textit{inliers ${\epsilon_i < \tilde{\epsilon}}$ are highly
    weighted}.}
    \renewcommand{\arraystretch}{1.15}
    \begin{tabularx}{\textwidth}{p{1.25cm}P{0.75cm}|ssss|bbb}
    \toprule
    \multicolumn{2}{c|}{\multirow{2}{*}{\secondupdate{Percent outliers}}} & Cauchy MAD &
    Tukey MAD & Welsch MAD & Var. Trimmed & Adaptive Barron~\cite{Barron2019} &
    Adaptive Chebrolu~\cite{Chebrolu2021} & Adaptive MB~(ours) \\
    \hline
    \multirow{4}{*}{\begin{tabular}{@{}c@{}} $\| \delta \mbshat{\phi} \|$ \\ $\left[ \deg \right]$ \end{tabular}}
        & \SI{20}{\percent} & 1.92 & 1.78 & 5.22 & 1.72 & 1.81-\colour{dark_grey}{2.36}-\textbf{\colour{light_grey}{2.91}} & 1.90-\colour{dark_grey}{2.39}-\colour{light_grey}{3.00} & \textbf{1.64}-\textbf{\colour{dark_grey}{2.23}}-\colour{light_grey}{2.95} \\ 
        & \SI{40}{\percent} & 1.80 & 1.72 & 2.15 & 1.76 & 1.86-\colour{dark_grey}{2.40}-\colour{light_grey}{3.09} & 1.91-\colour{dark_grey}{2.52}-\colour{light_grey}{3.27} & \textbf{1.68}-\textbf{\colour{dark_grey}{2.25}}-\textbf{\colour{light_grey}{2.84}} \\ 
        & \SI{60}{\percent} & 9.29 & 12.51 & 6.20 & 1.65 & 1.74-\colour{dark_grey}{2.32}-\colour{light_grey}{3.07} & 1.99-\colour{dark_grey}{2.58}-\colour{light_grey}{3.28} & \textbf{1.60}-\textbf{\colour{dark_grey}{2.23}}-\textbf{\colour{light_grey}{2.83}} \\ 
        & \SI{80}{\percent} & 5.88 & 8.89 & 2.42 & 1.95 & 2.11-\colour{dark_grey}{2.67}-\colour{light_grey}{3.28} & 2.88-\colour{dark_grey}{3.83}-\colour{light_grey}{4.96} & \textbf{1.78}-\textbf{\colour{dark_grey}{2.32}}-\textbf{\colour{light_grey}{2.84}} \\ 
    \hline
    \multirow{4}{*}{\begin{tabular}{@{}c@{}} $\| \delta \mbshat{\rho} \|$ \\ $\left[ \meter\!\meter \right]$ \end{tabular}}
        & \SI{20}{\percent} & 43 & \textbf{41} & 106 & 42 & 42-\colour{dark_grey}{64}-\textbf{\colour{light_grey}{77}} & 42-\colour{dark_grey}{67}-\textbf{\colour{light_grey}{77}} & 42-\textbf{\colour{dark_grey}{61}}-\colour{light_grey}{78} \\ 
        & \SI{40}{\percent} & 38 & 37 & 47 & \textbf{35} & 38-\colour{dark_grey}{66}-\colour{light_grey}{84} & 43-\colour{dark_grey}{65}-\colour{light_grey}{87} & 38-\textbf{\colour{dark_grey}{61}}-\textbf{\colour{light_grey}{80}} \\ 
        & \SI{60}{\percent} & 210 & 344 & 155 & \textbf{39} & 40-\colour{dark_grey}{61}-\colour{light_grey}{69} & 45-\colour{dark_grey}{65}-\colour{light_grey}{78} & \textbf{39}-\textbf{\colour{dark_grey}{53}}-\textbf{\colour{light_grey}{64}} \\ 
        & \SI{80}{\percent} & 139 & 277 & 52 & 39 & 46-\colour{dark_grey}{62}-\colour{light_grey}{80} & 67-\colour{dark_grey}{105}-\colour{light_grey}{131} & \textbf{36}-\textbf{\colour{dark_grey}{50}}-\textbf{\colour{light_grey}{64}} \\    
    \hline
        Time $\left[ \SI{}{\second} \right]$    & ALL & 0.04 & 0.04 & 0.04 & 0.02 & 0.18-\colour{dark_grey}{0.21}-\colour{light_grey}{0.23} & 0.22-\colour{dark_grey}{0.27}-\colour{light_grey}{0.37} & 0.13-\colour{dark_grey}{0.15}-\colour{light_grey}{0.17} \\
        Iterations & ALL & 7.0 & 8.0 & 7.0 & 3.0 & 6.0-\colour{dark_grey}{7.0}-\colour{light_grey}{7.0} & 8.0-\colour{dark_grey}{9.0}-\colour{light_grey}{12.0} & 4.0-\colour{dark_grey}{5.0}-\colour{light_grey}{5.0} \\
    \bottomrule
    \end{tabularx}
	\label{tab:pose_avg_table}
\end{table*}

\section{Conclusion}
\label{sec:conclusion}

Outlier rejection is a key component of real-world robotics problems.  Many
problems in robotics involve least-squares optimization, where the residual is
the norm of some multivariate, normally distributed error.  The residual will
then follow a Chi distribution, with a mode value of ${\tilde{\epsilon} > 0}$.
Existing RLFs assume a mode of zero, leading to the creation of a ``mode gap''
that impacts convergence times and optimization accuracy.  By accounting for
this gap, the proposed ``Adaptive MB'' approach is able to deliver faster
convergence times and more robust performance than existing adaptive RLFs
\cite{Barron2019,Chebrolu2021}, as well as the fixed RLFs studied.  This was
demonstrated for two fundamental state estimation problems, point cloud
alignment and pose averaging.  The results suggest this approach will be widely
applicable to least-squares optimization problems in state estimation and
robotics.

\section*{Acknowledgement}
\label{sec:acknowledgement}

\secondupdate{The authors would like to thank Mitchell Cohen for many insightful
discussions on the theory and application of robust loss functions.}

\printbibliography

\end{document}

%% file: z_arxiv-cover-ieee.tex
%
%
%
%
%
%
%
\def \myJournal {IEEE Robotics and Automation Letters}
\def \myDoi {10.1109/LRA.2022.3179424}
\def \myPaperSiteName {IEEE Xplore}
\def \myPaperSiteLink {https://ieeexplore.ieee.org/document/9786679}
\def \myYear {2022}
\def \myPaperCitation{T. Hitchcox and J. R. Forbes, ``Mind the Gap: Norm-Aware
Adaptive Robust Loss for Multivariate Least-Squares Problems,'' in \textit{IEEE
Robotics and Automation Letters}, vol. 7, no. 3, pp. 7116-7123, 2022.}


\begin{figure*}[t]

\thispagestyle{empty}
\begin{center}
\begin{minipage}{6in}
\centering
This paper has been accepted for publication in \emph{\myJournal}. 
\vspace{1em}

This is the author's version of an article that has, or will be, published in this journal or conference. Changes were, or will be, made to this version by the publisher prior to publication.
\vspace{2em}

\begin{tabular}{rl}
DOI: & \myDoi\\
\myPaperSiteName: & \texttt{\myPaperSiteLink}
\end{tabular}

\vspace{2em}
Please cite this paper as:

\myPaperCitation

\vspace{15cm}
\copyright \myYear \hspace{4pt}IEEE. Personal use of this material is permitted. Permission from IEEE must be obtained for all other uses, in any current or future media, including reprinting/republishing this material for advertising or promotional purposes, creating new collective works, for resale or redistribution to servers or lists, or reuse of any copyrighted component of this work in other works.

\end{minipage}
\end{center}
\end{figure*}
\newpage
\clearpage
\pagenumbering{arabic} 